\newif\ifarxiv \arxivtrue
\ifarxiv
\documentclass[acmtog,authorversion,nonacm]{acmart_arxiv/acmart}
\else
\documentclass[acmtog,anonymous,review]{acmart}
\acmSubmissionID{320}  %
\fi

\usepackage{booktabs} %
\usepackage{caption}
\usepackage{subcaption}
\usepackage[ruled]{algorithm2e} %

\SetAlFnt{\small}
\SetAlCapFnt{\small}
\SetAlCapNameFnt{\small}
\SetAlCapHSkip{0pt}

\acmJournal{TOG}

\newif\ifdraft \draftfalse
\newif\ifanonymous \anonymousfalse
\newif\ifappendix \appendixtrue

\usepackage{amsfonts}  %
\usepackage{soul}      %
\usepackage{cancel}      %
\usepackage{makecell}
\usepackage{colortbl}

\ifundef{\ifanonymous} {\newif\ifanonymous \anonymoustrue}  {}
\ifundef{\ifdraft}     {\newif\ifdraft \draftfalse}         {}
\ifundef{\ifarxiv}     {\newif\ifarxiv \arxivfalse}         {}

\ifdraft
\newcommand{\todo}[1]{{\color{red} #1}}
\newcommand{\gtc}[1]{\textcolor{cyan}{GT: #1}}

\newcommand{\gt}[1]{\textcolor{cyan}{#1}}
\newcommand{\gtdel}[1]{{\color{cyan} \st{#1}}}
\newcommand{\sr}[1]{{\color{violet} #1}}
\newcommand{\srr}[1]{{\color{red}\textbf{SR:} #1}}
\newcommand{\srv}[1]{{\color{violet}\textbf{SR:} #1}}

\newcommand{\dcc}[1]{{\color{red}\textbf{DC:} #1}}

\newcommand{\abc}[1]{{\color{purple}\textbf{AB:} #1}}
\else
\newcommand{\todo}[1]{}
\newcommand{\gtc}[1]{}
\newcommand{\gt}[1]{#1}
\newcommand{\gtdel}[1]{}
\newcommand{\sr}[1]{#1}
\newcommand{\srr}[1]{}
\newcommand{\srv}[1]{}

\newcommand{\dcc}[1]{}

\newcommand{\abc}[1]{}

\fi

\newcommand{\algoname}{{SinMDM}} 

\makeatletter
\ifundef{\onedot} 
{
\DeclareRobustCommand\onedot{\futurelet\@let@token\@onedot}
\def\@onedot{\ifx\@let@token.\else.\null\fi\xspace}
}
{}

\ifundef{\eg} {\def\eg{\emph{e.g}\onedot}} {} \ifundef{\Eg} {\def\Eg{\emph{E.g}\onedot}} {} 
\ifundef{\ie} {\def\ie{\emph{i.e}\onedot}} {} \ifundef{\Ie} {\def\Ie{\emph{I.e}\onedot}} {} 
\ifundef{\cf} {\def\cf{\emph{cf}\onedot}} {} \ifundef{\Cf} {\def\Cf{\emph{Cf}\onedot}} {} 
\ifundef{\etc} {\def\etc{\emph{etc}\onedot}} {} \ifundef{\vs} {\def\vs{\emph{vs}\onedot}} {} 
\ifundef{\wrt} {\def\wrt{w.r.t\onedot}} {} \ifundef{\dof} {\def\dof{d.o.f\onedot}} {} 
\ifundef{\iid} {\def\iid{i.i.d\onedot}} {}  \ifundef{\wolog} {\def\wolog{w.l.o.g\onedot}} {} 
\ifundef{\etal} {\def\etal{\emph{et al}\onedot}} {} 
\ifundef{\sup} {\def\sup{sup. mat\onedot}} {} 
\makeatother

\newcommand{\FK}{\text{FK}}

\newcommand{\bC}{{\bf C}}
\newcommand{\bD}{{\bf D}}

\newcommand{\bL}{{\bf L}}

\newcommand{\bS}{{\bf S}}

\newcommand{\dynfeat}{\bD}
\newcommand{\statfeat}{\bS}

\newcommand{\mll}{\mathcal{L}}

\newcommand{\nn}{\mathcal{N}}

\newcommand{\Loss}{\mll}

\newcommand{\R}{\mathbb{R}} %
\newcommand{\E}{\mathbb{E}} %
\newcommand{\indicator}{\mathbf{1}}

\newcommand{\norm}[1]{\lVert#1\rVert}

\newcommand*\rot{\rotatebox{90}}

\begin{document}
\title{Single Motion Diffusion}

\ifanonymous\else
\author{Sigal Raab}\authornote{equal contribution.}
\author{Inbal Leibovitch}\authornotemark[1]
\author{Guy Tevet}
\author{Moab Arar} 
\author{Amit H. Bermano}
\ifarxiv
\author{and Daniel Cohen-Or} 
\else
\author{Daniel Cohen-Or} 
\fi
\affiliation{%
  \institution{Tel-Aviv University}
  \country{Israel}
}
\email{sigalraab@tauex.tau.ac.il,{inball1,guytevet}@mail.tau.ac.il}

\renewcommand\shortauthors{Raab and Leibovitch \etal}
\fi

\ifarxiv
\else
\fi
\begin{abstract}

Synthesizing realistic animations of humans, animals, and even imaginary creatures, has long been a goal for artists and computer graphics professionals. 
Compared to the imaging domain, which is rich with large available datasets, the number of data instances for the motion domain is limited, particularly for the animation of animals and exotic creatures (\eg, dragons), which have unique skeletons and motion patterns.
In this work, we present a Single Motion Diffusion Model, dubbed \algoname{}, a model designed to 
learn the internal motifs of a single motion sequence with arbitrary topology and synthesize motions of arbitrary length that are faithful to them.
We harness the power of diffusion models and present a denoising network explicitly designed for the task of learning from a single input motion. 
\algoname{} is designed to be a lightweight architecture, which avoids overfitting by using a shallow network with local attention layers that narrow the receptive field and encourage motion diversity.
\algoname{} can be applied in various contexts, including spatial and temporal in-betweening, motion expansion, style transfer, and crowd animation. 
Our results show that \algoname{} outperforms existing methods both in quality and time-space efficiency.
Moreover, while current approaches require additional training for different applications, our work facilitates these applications at inference time.
Our code and trained models \ifanonymous{will be made available.}\else{are available at \url{https://sinmdm.github.io/SinMDM-page}.}\fi

\end{abstract}

\ifarxiv
\else


\fi

\maketitle

\section{Introduction} \label{sec:intro}
\begin{figure}
\centering
\includegraphics[width=\linewidth]{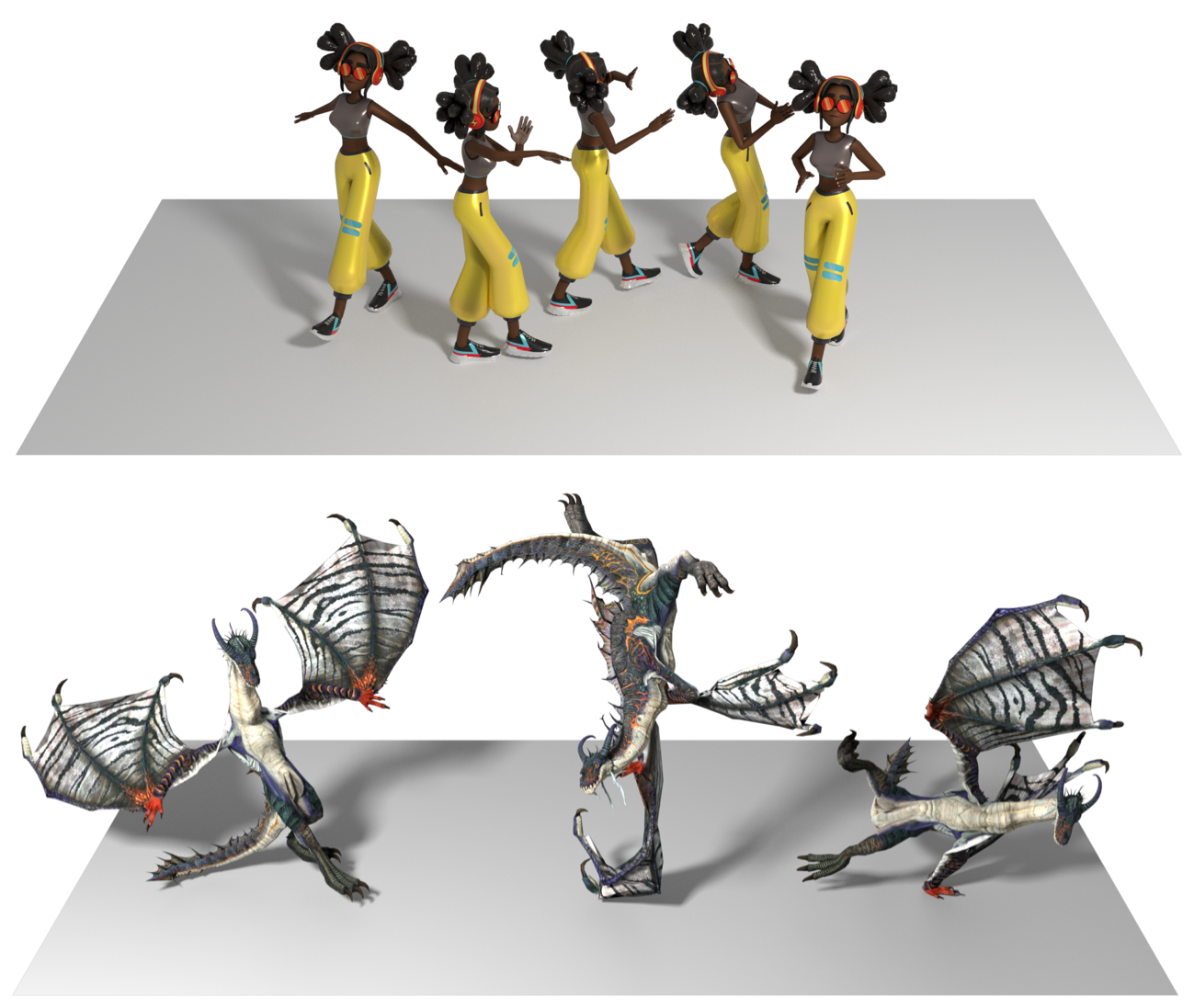}

\vspace{-10pt}
\caption{\algoname{} learns the internal motion motifs from a single motion sequence with arbitrary topology and synthesizes motions that are faithful to the learned core motifs of the input sequence. Top: a girl exercising while walking.
Bottom: a breakdancing dragon. Left to right: breakdance uprock, breakdance freeze, and breakdance flair.  }
\label{fig:teaser}
\vspace{-5pt}
\end{figure}

3D character animation is a long pursued task in computer graphics with many applications, from the big screen to virtual reality headsets. It is notoriously known as a time-consuming task done by expert artists.
In recent years, neural models have offered faster and less expensive tools for modeling motion \cite{holden2016deep,petrovich2022temos,raab2022modi}. In particular, the very recent adaptation of diffusion models into the motion domain provides unprecedented results in both quality and diversity \cite{tevet2022human,kim2022flame}.

These data-driven methods typically require large amounts of data for training. 
However, motion data is quite scarce and, moreover, for a non-human skeleton, it is barely existent.
The few available datasets contain humanoids only, whose topology and bone ratio are fixed.
Animators often customize a skeleton per character (human, animal, or magical creature), for which common data-driven techniques are irrelevant. 

In this work, we present a Single Motion Diffusion Model, dubbed \algoname{}, that trains on a \textit{single motion} input sequence.
Our model enables modeling motions of arbitrary skeletal topology, which often have no more than one animation sequence to learn from. 
\algoname{} can synthesize a variety of variable-length motion sequences that retain the core motion elements of the input
and can handle complex and non-trivial skeletons. 
For example, our model can generate a diverse clan based on one flying dragon or one hopping ostrich.

Learning from a single instance has been explored for the imaging domain~\cite{shaham2019singan,shocher2019ingan} and for the motion domain~\cite{li2022ganimator}, using the GAN architecture~\cite{goodfellow2014generative}. 
Indeed until recently, GANs have been the dominant approach for generative models. 
We find diffusion models ~\cite{ho2020denoising} more suitable for single input learning, as the descriptive ability attained by gradual denoising  yields a lightweight model that, compared to prior art, is simpler in architecture and more efficient in terms of the number of parameters and training time.
Furthermore, we demonstrate that diffusion models can be effectively utilized with limited data, challenging the notion
that they solely rely on large datasets.

To learn local motion sequences, the receptive field must be small enough, analogously to the use of patch-based discriminators~\cite{isola2017image,li2016precomputed} in GAN-based techniques. The use of a narrow receptive field (Fig.~\ref{fig:architecture}) promotes diversity and reduces overfitting. We show the importance of narrow receptive fields in our ablation studies.

Most motion diffusion models use transformers. However, vanilla transformers are not suitable for learning a single sequence, as their receptive field encompasses the entire motion. A similar challenge exists in the UNet architecture, which is common for image diffusion models. Its depth, combined with global attention layers, induces a receptive field that covers the whole motion.

\algoname{} leverages the concept of narrow receptive fields and introduces a motion architecture specifically designed with this concept.
It combines a shallow UNet~\cite{ronneberger2015u} model adapted for motion with a QnA~\cite{arar2022learned} local attention mechanism, instead of global attention. 
As a result, \algoname{} outperforms prior art both quantitatively and qualitatively, and demonstrates high efficiency with shorter training time and less memory consumption. Imputed to its lightweight architecture, \algoname{} can be trained on a single mid-range GPU.

We present many use cases of \algoname{}. 
While prior works require designated training per application, ours are applied at inference time, with no need to re-train.
Moreover, applications that
require different dedicated algorithms in prior art,
are here grouped together as special cases of the same technique, significantly simplifying their use.
One of the applications we present is \emph{Motion Composition}, where a given motion sequence is composed jointly with a synthesized one, either temporally or spatially. Special cases of motion composition include in-betweening and motion expansion.
Another application that we present
is \emph{Harmonization}, along with its special case, style transfer. Here, a reference motion is modified to match the learned motion motifs. It should be emphasized that implementing style transfer using a denoising model is a non-trivial task, and enabling it through motion harmonization is unique.
We further present two more applications: \emph{long sequence generation} and \emph{crowd animation}.

In our presented work, we suggest two comprehensive benchmarks for single-motion evaluation. 
The first is built upon the artistically crafted MIXAMO \shortcite{mixamo} dataset, utilizing metrics that do not require an additional feature-extracting model. The second is based on the HumanML3D \shortcite{guo2022generating} dataset and enables metrics that use latent features, such as single-FID. 
We show that our model outperforms current works on both benchmarks.

\ifarxiv\else
\vspace{-5pt}
\fi
\section{Related Work} \label{sec:related_work}

\subsection{Single-Instance Learning}

The goal of single-instance generation is to learn an unconditional generative model from a single instance and generate diverse samples with similar content by capturing the internal statistics of patches.
The type of instance depends on the input domain.
Most single-instance learning research has been focused on the domain of imaging. The first works on this topic are SinGAN~\cite{shaham2019singan} and InGAN~\cite{shocher2019ingan}. SinGAN uses a patch-based discriminator~\cite{isola2017image,li2016precomputed} and an image pyramid to generate diverse results hierarchically. InGAN~\cite{shocher2019ingan}, uses a conditional GAN to solve the same problem using geometry transformation. More recent approaches include ExSinGAN~\cite{zhang2021exsingan}, which trains multiple modular GANs to model the distribution of structure, semantics, and texture, and ConSinGAN~\cite{hinz2021improved}, which trains several stages sequentially and improves SinGAN. Many works in the imaging domain follow and improve the aforementioned pioneering works~\cite{asano2020acritical,granot2022drop,chen2021mogan,lin2020tuigan,sun2020esingan,sushko2021generating,yoo2021sinir,zhang2022petsgan,zheng2021patchwise}. 

Several works have been introduced in other domains, such as for the shapes domain~\cite{son2022singraf} and for the 3D scenes domain~\cite{son2022singraf}.
In the motion domain, the only work that learns a single motion is Ganimator~\cite{li2022ganimator}. Ganimator follows SinGAN, hence it uses a GAN architecture, with a patch-based discriminator and a temporal pyramid.

The vast majority of single-instance learning works use a GAN architecture~\cite{goodfellow2014generative}. Until recently, GANs have been the dominant approach for generative models. However, we are currently seeing a trend towards using diffusion models~\cite{ho2020denoising, song2020denoising} as an alternative to GANs.

A number of concurrent works in the imaging domain use diffusion models to learn from single images. 
Similar to our approach,
\citet{wang2022sindiffusion} and \citet{nikankin2022sinfusion} drop the image pyramid structure and use a UNet~\cite{ronneberger2015u} with limited depth. A different work~\cite{kulikov2022sinddm} constructs a multi-scale diffusion process from down-sampled versions of the training image, as well as their blurry versions.

Ganimator~\cite{li2022ganimator} is our immediate comparison reference, as it is the only single-motion learning work. Sec. \ref{sec:experiments} and our supplementary video show that \algoname{} outperforms it quantitatively and qualitatively. In addition, Ganimator uses a complex architecture that combines a temporal hierarchy of motions with a skeletal hierarchy of joints.
Our model uses neither hierarchies, which makes it simple in architecture and efficient in time and space,
while achieving better results. 

\ifarxiv\else
\vspace{-5pt}
\fi
\subsection{Diffusion Models}

Diffusion models use a stochastic diffusion process, as modeled in thermodynamics~\cite{sohl2015deep, song2020improved}, to generate samples from a data distribution. 
These models are adapted for image generative applications. \citet{dhariwal2021diffusion} introduce the concept of classifier-guided diffusion for conditional generation, which is later adapted in the GLIDE~\cite{nichol2021glide} model. %
\citet{ho2022classifier} propose the Classifier-Free Guidance approach,  %
which can trade-off between fidelity and diversity in the generated samples. This approach has been demonstrated to achieve better results compared to other methods, as shown by \citet{nichol2021glide}.

Local editing of images may be viewed as an inpainting problem, in which a portion of the image is held constant while the model denoises the remaining part \cite{song2020score, saharia2022palette}. %
In our work, we adapt this technique for motion composition of specific body parts or temporal intervals. 

In the motion domain, several very recent works~\cite{tevet2022human,zhang2022motiondiffuse,kim2022flame} introduce diffusion-based synthesis, where the most prominent one is MDM~\cite{tevet2022human}. MDM utilizes a lightweight network,
uses a transformer rather than the common UNet and predicts motion rather than noise.
Its general design has already been used for various motion applications~\cite{shafir2023human,tseng2022edge,yuan2022physdiff}.
Like MDM, \algoname{} presents a lightweight architecture and predicts motion rather than noise. %
However, unlike MDM, our work uses a QnA-based UNet rather than a transformer, as the receptive field of a transformer is the full motion, inducing over-fitting.

\ifarxiv\else
\vspace{-5pt}
\fi
\subsection{Motion Synthesis Models}

In recent years, we witness prosperity in the domain of motion synthesis using neural networks~\cite{holden2015learning, holden2016deep}. Most of these models focus on specific motion-related tasks, conditioned on some limiting factors, which can be high-level guidance such as action~\cite{petrovich2021actor,guo2020action2motion,cervantes2022implicit, tevet2022human} or text~\cite{tevet2022motionclip,zhang2021write, petrovich2022temos,ahuja2019language2pose,guo2022generating,bhattacharya2021text2gestures, tevet2022human},
can be parts of a motion such as motion prefix ~\cite{aksan2019structured,barsoum2018hp,habibie2017recurrent,yuan2020dlow,zhang2021we,hernandez2019human} or in-betweening \cite{harvey2020robust,duan2021single,kaufmann2020convolutional,harvey2018recurrent}, motion retargeting or style transfer~\cite{holden2017fast,villegas2018neural,aberman2019learning, aberman2020skeleton,aberman2020unpaired}, and even music ~\cite{aristidou2022rhythm,sun2020deepdance,li2021learn,lee2018listen}. Fewer models are fully unconditioned~\cite{holden2016deep,raab2022modi,starke2022deepphase} and they learn the motion manifold from the input data in an unsupervised manner.

The architecture of motion synthesis models can be roughly divided into recurrent~\cite{fragkiadaki2015recurrent,zhou2018auto,habibie2017recurrent,ghorbani2020b},  auto encoder based~\cite{maheshwari2022mugl,guo2020action2motion,Jang2020constructing, petrovich2021actor}, GAN based ~\cite{degardin2022generative, wang2020learning, yan2019convolutional,yu2020structure}, normalizing flows based~\cite{henter2020moglow}, and more recently, neural field based~\cite{he2022nemf} and diffusion based~\cite{tevet2022human,zhang2022motiondiffuse,kim2022flame}.
Our work belongs to the latter category.
\ifarxiv
\section{Preliminary} 
\else
\ifarxiv\else
\vspace{-5pt}
\fi
\section{Preliminary} %
\fi

In this work, we present \algoname{}, a novel framework that learns the internal motion motifs of a \textit{single motion} of arbitrary topology, and generates a variety of synthesized motions that retain the core motion elements of the input sequence.

At the crux of our approach lays a denoising diffusion probabilistic model (DDPM) \cite{ho2020denoising}.
We consider diffusion models to be more appropriate for single input learning compared to previous methods and suggest a lightweight model, efficient in time and space and simple in architecture. This is achieved through the gradual denoising process, which enhances the model's descriptive capability.
Our generative network is a UNet~\cite{ronneberger2015u} whose attention layers are replaced by the recently introduced QnA layers \cite{arar2022learned}. 

In the rest of this section, we briefly recap DDPM and describe our motion representation. In the following section, we describe our method and focus on our design choices. 
Next, we describe various applications enabled by \algoname{} (Sec.~\ref{sec:apps}), detail the experiments conducted to validate our approach (Sec.~\ref{sec:experiments}), and summarise with conclusions (Sec.~\ref{sec:conclusion}). The readers are encouraged to watch the supplementary video in order to get a full impression of our results.

\vspace{-5pt}
\subsection{Denoising Diffusion Probabilistic Models (DDPM)} \label{sec:ddpm}
DDPMs~\shortcite{ho2020denoising} have become the de-facto leading generative networks technique. While they have primarily dominated the imaging domain \cite{dhariwal2021diffusion}, recent works have successfully applied this approach in the motion domain \cite{tevet2022human,zhang2022motiondiffuse}. 
Denoising networks learn to convert unstructured noise to samples from a given distribution, through an iterative process of progressively removing small amounts of Gaussian noise. 

Given an input motion sequence $x_0$, we apply a Markov noising process of $T$ steps, $\{x_t\}_{t=0}^T$,  such that 
\begin{equation}
{
\ifarxiv\else
\fi
q(x_{t} | x_{t-1}) = \nn(\sqrt{\alpha_{t}}x_{t-1},(1-\alpha_{t})I),
\ifarxiv\else
\fi
}
\vspace{-5pt}
\end{equation}
where $\alpha_{t} \in (0,1)$ are constant hyper-parameters. When  $\alpha_{t}$ is small enough, we can approximate $x_{T} \sim \nn(0,I)$.

We apply unconditional motion synthesis that models $x_0$ as the reversed diffusion process of gradually cleaning $x_{T}$, using a generative network $p_\theta$. Following \citet{tevet2022human} we choose to predict the input motion, denoted $\hat{x}_{0}$ \cite{ramesh2022hierarchical} rather than predicting $\epsilon_t$, hence  
\begin{equation} \label{eq:p_theta}
\ifarxiv\else
\fi
\hat{x}_0 = p_\theta(x_t, t). 
\end{equation}

We apply the widespread diffusion loss, via

\begin{equation}
\Loss_\text{simple} = \E_{t \sim [1,T]}\norm{ x_0 - p_\theta(x_t, t)}_2^2.
\end{equation}

Synthesis at inference time is applied through a series of iterations, starting with pure noise $x_T$. 
In each iteration, a clean version of the current sample $x_t$ is predicted using a generator $p_\theta$. This predicted clean sample $\hat{x}_{0}$ is then "re-noised" to create the next sample $x_{t-1}$, with the process being repeated until $t=0$ is reached.

\ifarxiv\else
\vspace{-5pt}
\fi
\subsection{Motion representation} \label{sec:repr}

A motion sequence is represented by its dynamic and static features, $\dynfeat$ and $\statfeat$, respectively. The former differ at each temporal frame (\eg, joint rotation angles), while the latter is temporally fixed (\eg., bone lengths). $\dynfeat$ and $\statfeat$ can be combined into global 3D pose sequences using \emph{forward kinematics} (\FK).
In our research, we focus on synthesizing 
the \emph{dynamic features}, leaving the static features intact. 
That is, we predict dynamics for a fixed skeleton topology with fixed bone lengths.
For simplicity, we use the term \emph{motion synthesis} for the generation of dynamic features only.

Let $N$ denote the number of frames in a motion sequence, and $F$ denote the length of the features describing a single frame. In the HumanML3D dataset, a frame is redundantly represented with the root position and joint positions, angles, velocities, and foot contact~\cite{guo2022generating}. For the other datasets used in this work, a frame is represented by  joint angles, root positions, and foot contact labels. 
We represent the dynamic features of a motion by a tensor $\dynfeat \in \R^{N\times F}$.
Naturally, the convolution for this representation is 1D, convolving on the temporal dimension (of size $N$) and holding $F$ features. 
\ifarxiv
Let $J$ denote the number of skeletal joints, and let $Q$ denote the number of rotational features, where rotational features may be Euler angles, quaternions, 6D rotations, etc. 
Let $C$ denote the number of joints that are prone to contact the ground.
Clearly, a human, a spider, and a snake possess different values of $C$.

When using the HumanML3D \cite{guo2022generating} dataset, we adhere to its representation, in which a single pose is defined by 
\[
p = (\dot{r}^{a}, \dot{r}^x, \dot{r}^z, r^y, j^p, j^v, j^r, c^f) \in \R^F,
\]
where $\dot{r}^{a} \in \R$ is the root angular velocity along the Y-axis. $\dot{r}^x, \dot{r}^z \in \R$ are root linear velocities on the XZ-plane, and $r^y \in \R$ is the root height. $j^p \in \R^{3J}$, $j^v \in \R^{3J}$ and $j^r \in \R^{6J}$ are the local joint positions, velocities, and rotations with respect to the root, and $c^f \in \R^4$ are binary features denoting the foot contact labels for four foot joints (two for each leg).

When using data from other datasets, we adhere to the representation used by Ganimator \shortcite{li2022ganimator}, so we can conduct a fair comparison with their results. Their representation consists of a 3D root location, a rotation angle for each joint, and foot contact labels. Altogether, for a general representation $D\in\R^{N\times F}$, we have got $F=3 + JQ + C$.

The rotations in both representations are defined in the coordinate frame of their parent in the kinematic chain, and are represented by the 6D rotation features ($Q = 6$) \citet{zhou2019continuity}, which yields the best result in many works \cite{qin2022motion,petrovich2021actor}.

A growing number of works use foot contact labels \cite{gordon2022flex, raab2022modi} to mitigate common foot sliding artifacts. 
Let $\bC$ denote the set of joints that contact the ground in the subject whose motion is being learned such that $C=|\bC|$.
The foot contact labels are represented by $\bL\in\{0,1\}^{N\times C}$. 

When a dataset provides foot contact label information \cite{guo2022generating}, we use it as is. When a dataset does not provide them, we calculate it as done by \citet{li2022ganimator}, via

\begin{equation}
\forall j\in \bC,n\in [1,N]: \quad\quad \bL^{nj} = \indicator[\norm{\Delta_n\FK(\left[\dynfeat, \statfeat\right])^{nj}}_2 < \epsilon],
\end{equation}

where $\Delta_n\FK(\left[\dynfeat, \statfeat\right])^{nj}$ denotes the velocity of joint $j$ in frame $n$ retrieved by a forward kinematics operator, and $\indicator[\cdot]$ is an indicator function. 

\else
More motion representation details can be found in our sup. mat.
\fi

\ifarxiv\else
\fi
\section{Generative network}

\begin{figure*}
\centering
\begin{subfigure}[b]{.37\linewidth}  
     \centering
     \includegraphics[height=3.6cm]{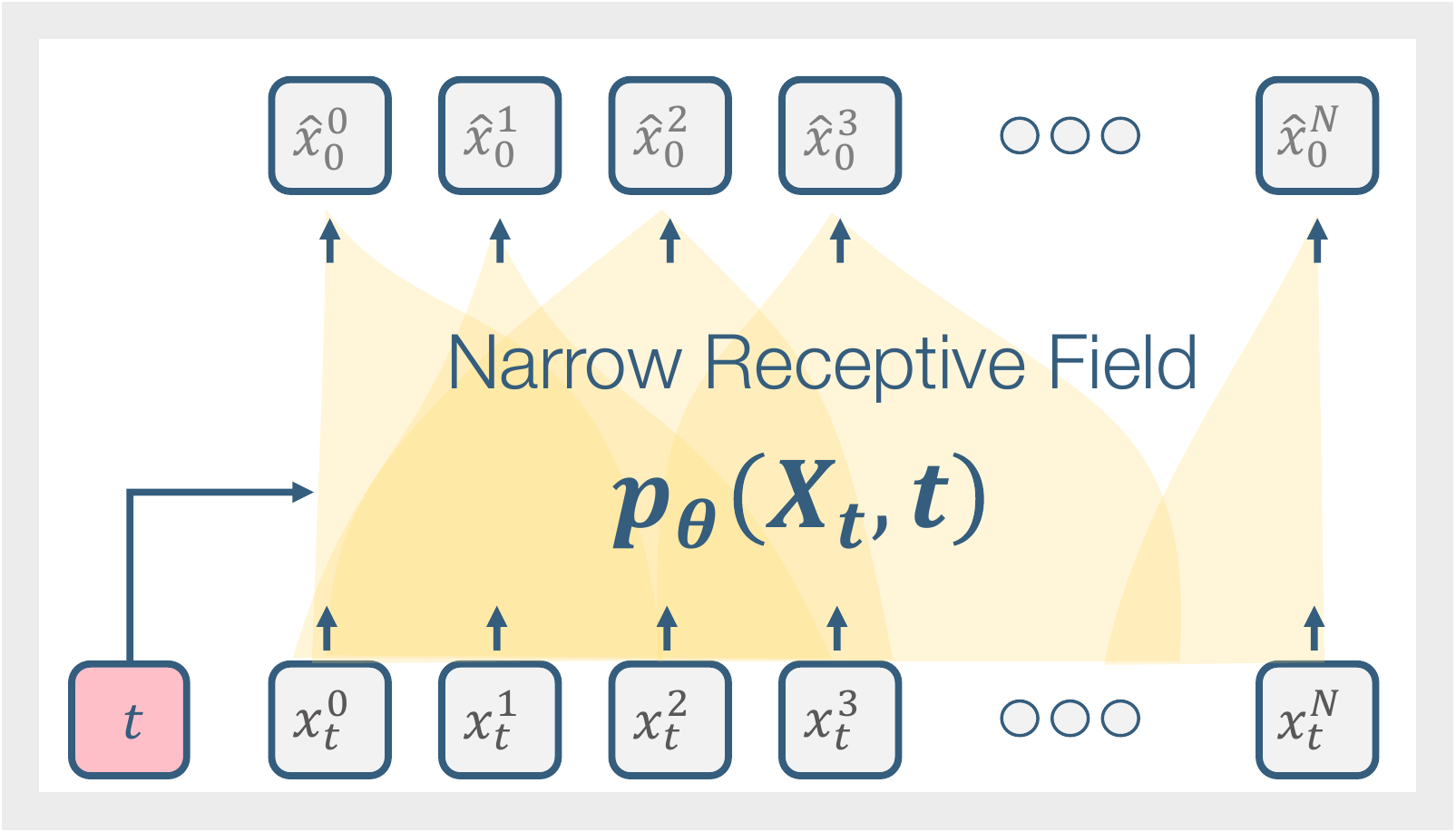} 
\end{subfigure}
\hfill
\begin{subfigure}[b]{.62\linewidth} 
     \centering
     \includegraphics[height=3.6cm]{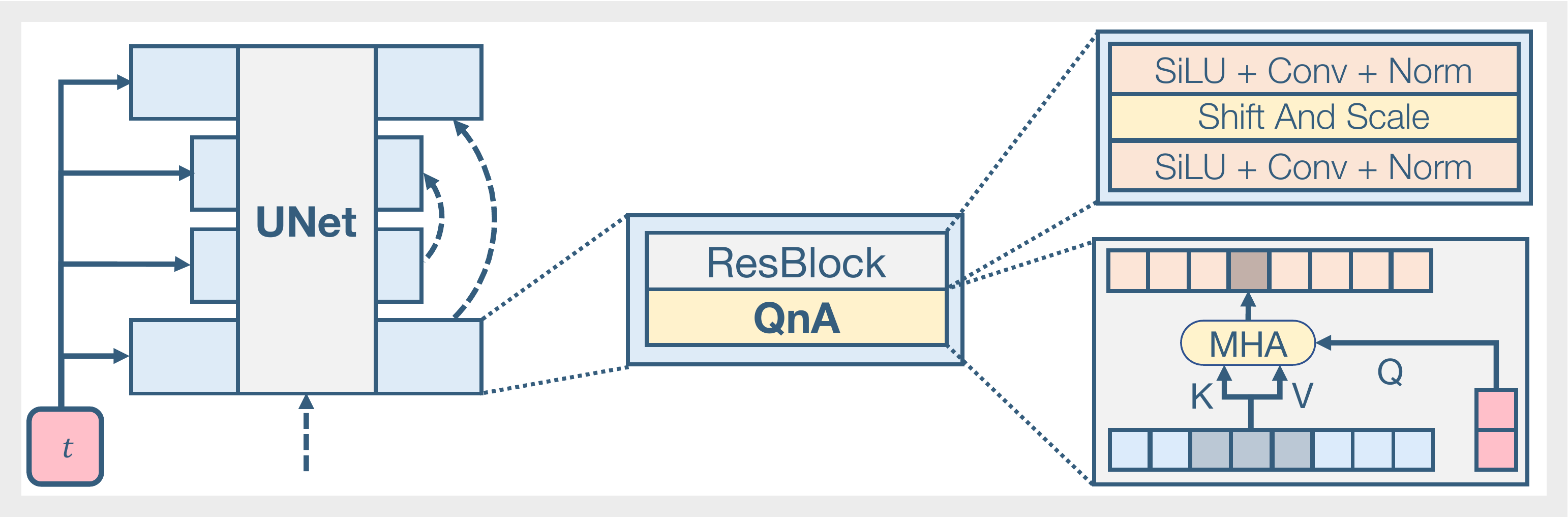}
\end{subfigure}

\vspace{-10pt}
\caption{\sr{
Left: To allow training on a single motion, our denoising network is designed such that its overall receptive field covers only a portion of the input sequence. This effectively allows the network to simultaneously learn from multiple local temporal motion segments. Our denoiser predicts the input sequence from a noisy one. $x_t^0\dots x_t^N$ is a motion of $N$ frames at diffusion step $t$. Right: Our network is a shallow UNet, enhanced with a QnA local attention layer.}
}
\label{fig:architecture}
\vspace{-5pt}
\end{figure*}

Our goal is to construct a model that can generate a variety of synthesized motions that retain the core motion motifs of a single learned input sequence.
More formally, we would like to construct the generative network  $p_\theta$ ( Eq.~\ref{eq:p_theta}) that synthesizes a motion $\hat{x}_0$ from a noised motion $x_t$.

While traditional single-instance techniques use a pyramid of down-sampled instances (images or motions) and learn in a coarse-to-fine fashion, our model introduces a simple architecture that requires no pyramids.

Our key insight is that internal motifs are learned more effectively with a limited receptive field (Fig.~\ref{fig:architecture} Left). We design \algoname{}, a novel generative architecture, accordingly.
Our model is a QnA-based degenerated UNet (Fig.~\ref{fig:architecture} Right).
The UNet architecture~\cite{ronneberger2015u} is frequently used by diffusion models in the imaging domain~\cite{nichol2021glide}. 
However, training a UNet on a single input leads to significant overfitting due to its large receptive field, resulting in synthesized sequences that closely resemble the input.

Our first design choice 
in mitigating this issue 
is to decrease the depth of the UNet and thereby limit the receptive field width.
However, this step alone is not enough, since standard UNets employ global attention layers, resulting in a receptive field that encompasses the entire sequence. 
A possible solution would be using local attention in non-overlapping windows, like in ViT \cite{dosovitskiy2021animage}. Nonetheless, non-interleaving windows tend to limit the cross-window interaction, compromising the model's performance.
Our solution is to use QnA \cite{arar2022learned}, a state-of-the-art shift-invariant local attention layer, that aggregates the input locally in an overlapping manner, much like convolutions, but with the expressive power of attention. The key idea behind QnA is to introduce learned queries, shared by all windows, allowing fast and efficient implementation. In particular, QnA enables local attention with a temporally narrow receptive field.
Our QnA-based UNet is the first to be used in the motion domain,
where we plug QnA layers instead of the global attention layers of a vanilla UNet.
\sr{QnA is substantially more efficient than global attention in terms of space and time, and our model benefits from this advantage as a byproduct.}
A detailed description of the QnA layers is available in \ifappendix{Appendix \ref{app:qna}.}\else{the sup. mat.}\fi 

In Sec.~\ref{sec:experiments}, we validate these design choices. \sr{We show the effectiveness of a narrow receptive field, and justify the usage of QnA layers and the choice of a UNet rather than a transformer.}

In \ifappendix{Appendix~\ref{app:hyper_params}}\else{our supplementary material}\fi, we provide a comprehensive list of the hyperparameters that can be used to reconstruct our results.

\ifarxiv\else
\vspace{-5pt}
\fi

\section{Applications} \label{sec:apps}

Single-motion learning using diffusion models enables various applications. 
All our applications are applied at inference time, with no need to re-train. 
This is in contrast to the only current single-motion synthesis work, Ganimator~\shortcite{li2022ganimator}, which requires specialized training for most of its applications.
To meet paper length limits and given the variety of potential applications, we illustrate six selected ones.
\sr{Note that applications that require different dedicated algorithms in prior art, are grouped together here as special cases of the same technique, significantly simplifying their use.}

In the following, we show \emph{Motion Composition}~(Sec. \ref{par:motion_comp}), where a given motion sequence is composed jointly with a synthesized one, either temporally or spatially. Special cases of motion composition include \emph{in-betweening}, \emph{motion expansion}, \emph{trajectory control}, and \emph{joints control}. 
With our \emph{Motion Harmonization}~(Sec. \ref{par:harmonization}), a reference input motion is altered to align with the learned motion motifs. We illustrate one important special case, \emph{style transfer}.
Lastly, we show how straightforward use~(Sec. \ref{par:straightforward}) of \algoname{} enables \emph{one shot long motion generation} and \emph{crowd animation}.
The applications presented here are also demonstrated in our supplementary video.

\ifarxiv\else
\vspace{-5pt}
\fi

\subsection{Motion Composition}
\label{par:motion_comp}
\ifarxiv
\begin{figure}%
\centering
\includegraphics[width=\linewidth]{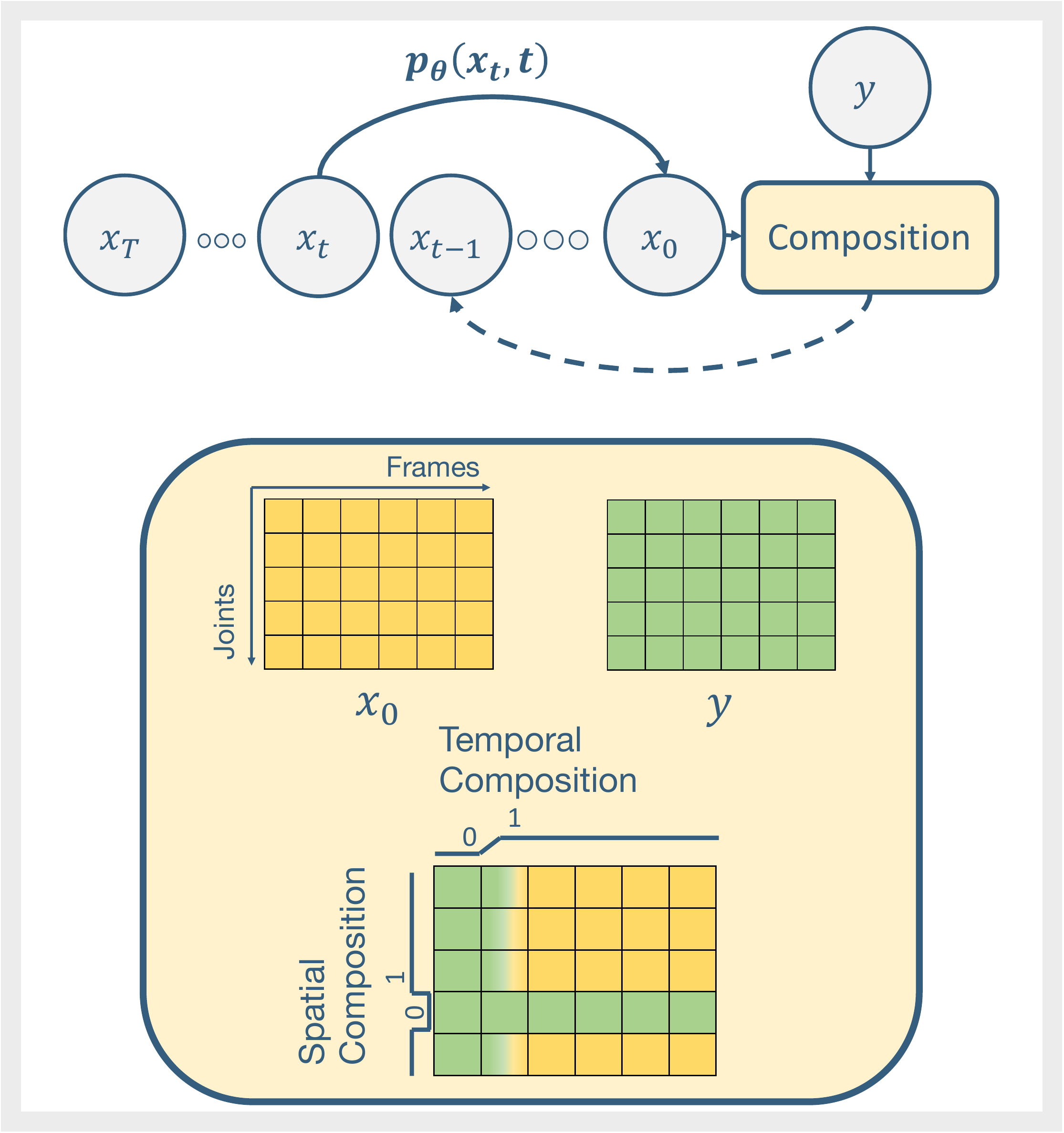} 

\vspace{-5pt}

\caption{Motion composition. Parts from a reference motion $y$, are composed with the synthesized motion $\hat{x}_0$, according to a composition map. }
\label{fig:composition_mask}
\end{figure}
\fi
Given a reference motion sequence $y$, and a region of interest (ROI) mask $m$, our goal is to synthesize a new motion $\hat{x}_0$, such that the regions of interest $\hat{x}_0\odot m$ are synthesized from random noise, while the complementary area remains as close as possible to the given motion $y$, \ie, $y \odot (1 - m) \approx \hat{x}_0 \odot (1 - m)$, where $\odot$ is element-wise multiplication. 
The model should output a coherent motion sequence, where the transition between given and synthesized parts is seamless. Moreover, the reference motion can be an arbitrary one, on which our model has \emph{not} been trained.

When using a binary mask \cite{avrahami2022blended}, as the reference motion  $y$ deviates from the motion the model was trained on, the blending between the given and synthesized parts becomes less smooth.
To mitigate this issue, we change the ROI mask such that the borders between the given and the synthesized motion segments are linearly interpolated, as depicted in Fig.~\ref{fig:composition_mask}.

We fix the motion segments that need to remain unchanged and sample the parts that need to be filled in. 
Each step of the iterative inference process (described in Sec.~\ref{sec:ddpm}) is slightly changed, such that parts of $y$ are assigned into $\hat{x}_0$ according to the indices of the mask. That is, $\hat{x}_0 \odot (1 - m) \leftarrow y \odot (1 - m)$. 

\ifarxiv\else
\vspace{-5pt}
\fi

\paragraph{Temporal composition -- use cases: in-betweening, motion expansion}
\ifarxiv
\ifarxiv 
\begin{figure}
\else 
\begin{figure} 
\fi

\centering

\includegraphics[width=\columnwidth]{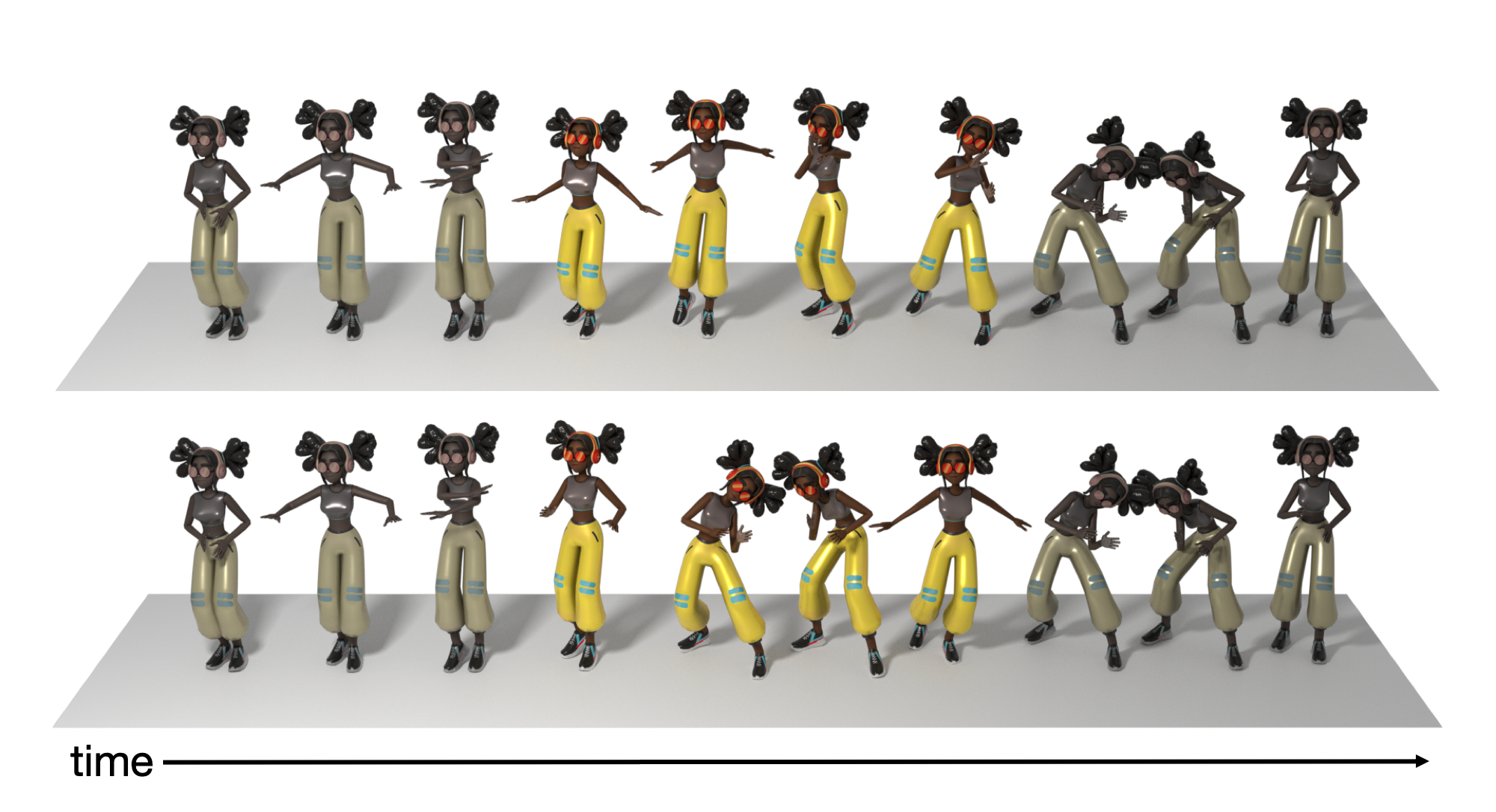}

\vspace{-10pt}
\caption{Temporal composition -- In-betweening. Both top and bottom show results for the same input, introducing diverse outputs. The beginning and the end of the motion are given by the reference sequence and can be distinguished according to their faded tone. Observe that the beginning and the end are identical in both sequences. The center of each motion is synthesized.  }
\label{fig:temporal_comp_inbetween}

\ifarxiv 
\end{figure}
\else 
\end{figure} 
\fi

\fi
\begin{figure}
\centering
    \begin{subfigure}{\linewidth}
    \includegraphics[width=\linewidth]{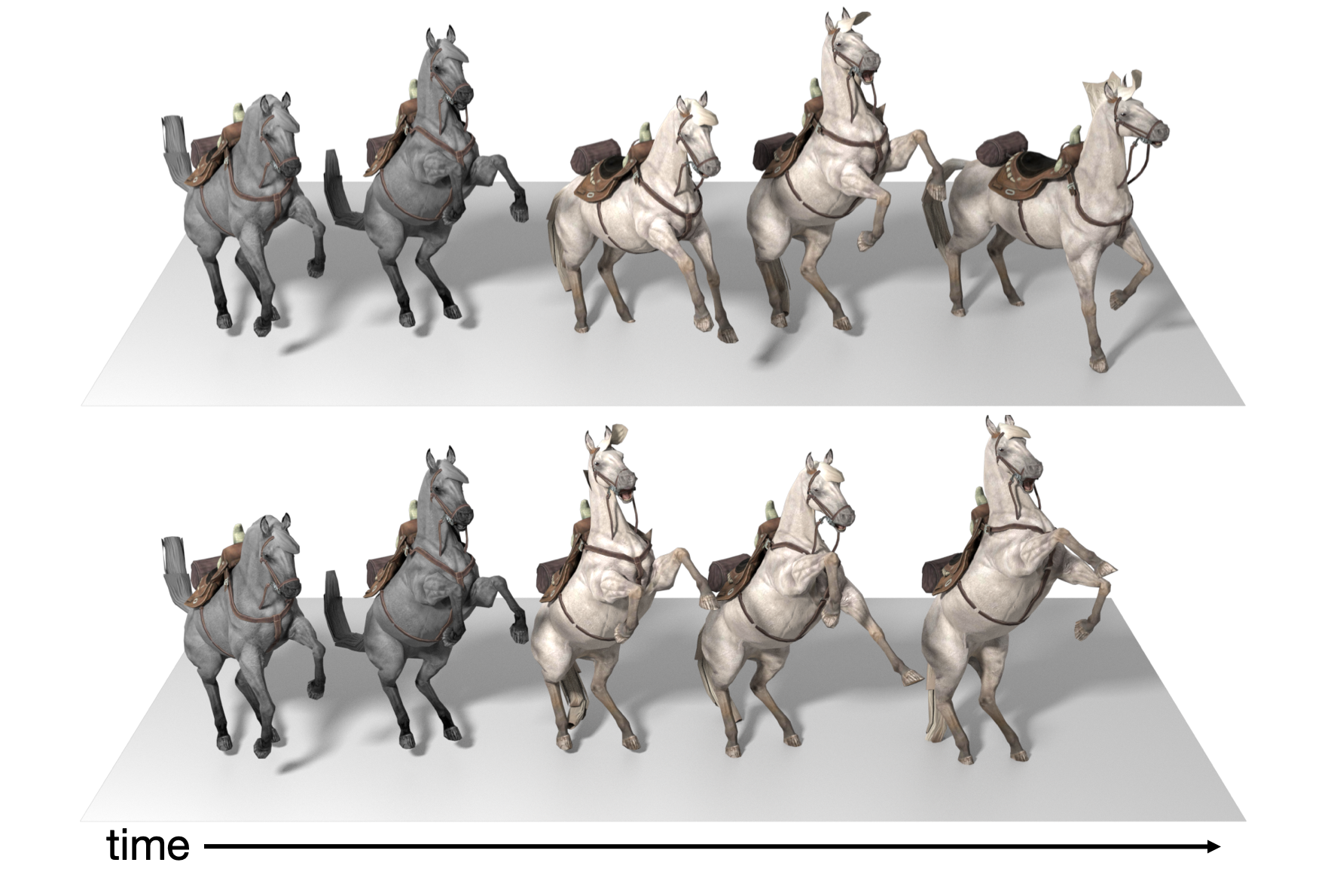}

    \end{subfigure}
    
    \begin{subfigure}{\linewidth}
    \includegraphics[width=\linewidth]{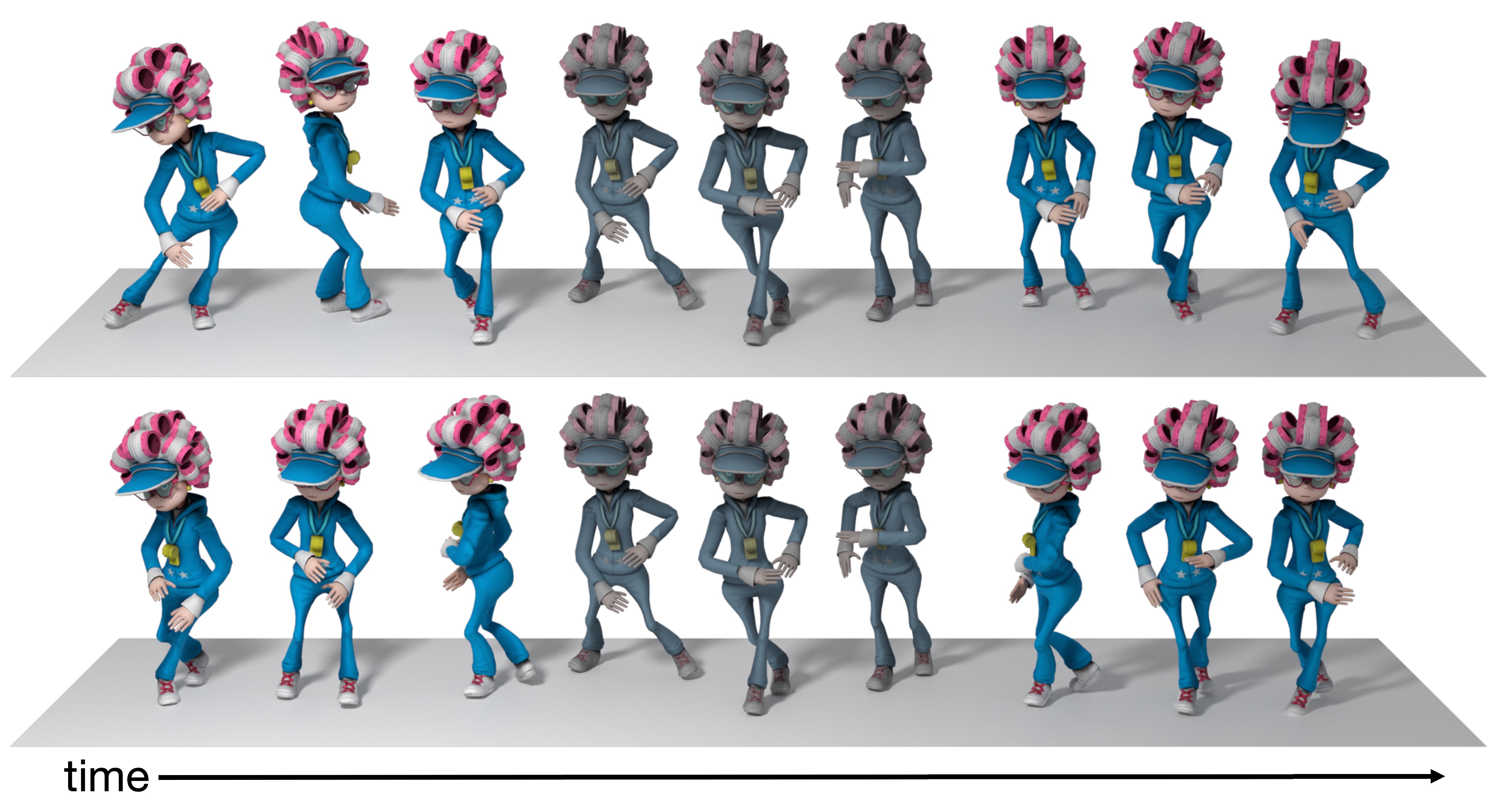}

    \vspace{-5pt}
    \end{subfigure}

\vspace{-5pt}
\caption{Temporal composition -- motion expansion. 
Pairs of motions exhibit diverse synthesis from a single input. 
The motion part provided by the reference sequence is identifiable by its faded color. 
Note that the parts given as input are identical in both sequences, 
while the synthesized parts differ. 
Top: synthesize a suffix given a temporal prefix. Bottom: synthesize a prefix and a suffix, given the middle part.}
   
\label{fig:temporal_comp_expansion}
\vspace{-5pt}
\end{figure}
Temporal composition is the action of filling in selected frame sequences. 
\emph{In-betweeining} \cite{harvey2020robust}, depicted in Fig.~\ref{fig:temporal_comp_inbetween}, is a special case of temporal composition, where the filled-in part is at the temporal interior of the sequence, and the reference $y$ is from the same distribution as the learned motion. 
Another special case of temporal composition is \emph{motion expansion}, the motion domain's equivalent of image outpainting \cite{yu2019free,lin2021infinitygan,teterwak2019boundless}, where the model generates content that resides beyond the edges of a reference motion sequence. In the case of motion expansion, the ROI mask is zeroed in the center frames, and assigned ones in the outer regions. See Fig.~\ref{fig:temporal_comp_expansion}.

\ifarxiv\else
\vspace{-5pt}
\fi

\paragraph{Spatial composition -- use cases: trajectory control, joints control}
\ifarxiv
\begin{figure}%
\centering
\includegraphics[width=\linewidth]{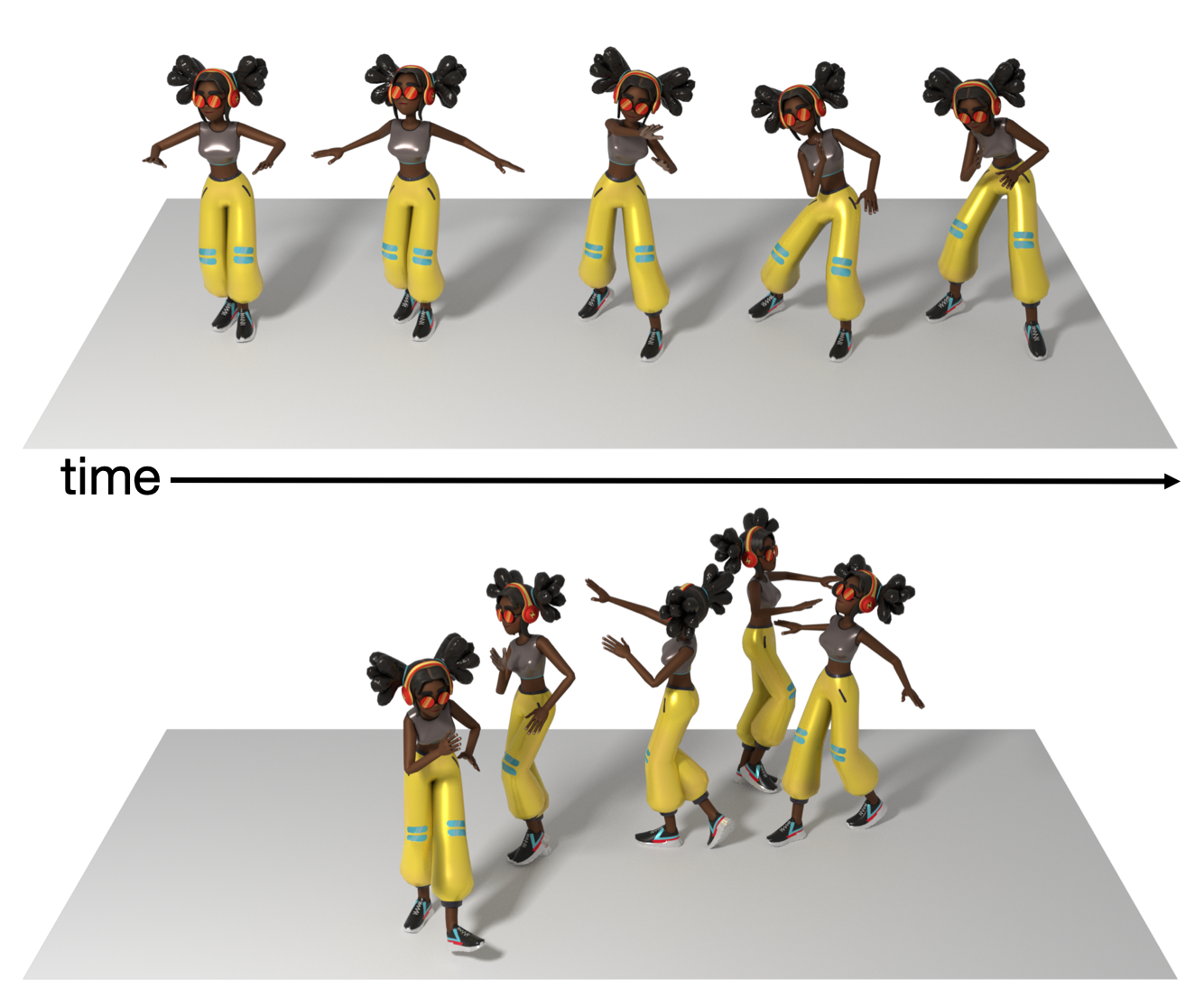} 

\vspace{-10pt}

\caption{Spatial composition. Top: reference motion, unseen by the network. Bottom: composed motion. 
The referenced sequence is \emph{warm-up}, and the learned one is \emph{walk in circle}. In the composed result, the top body part performs a warm-up activity, and the bottom body part walks in a curved path.
}
\label{fig:spatial_comp}
\end{figure}
\fi
Motion composition can be applied spatially, by assigning selected joint indices to the ROI mask.
In Fig.~\ref{fig:spatial_comp} we illustrate control over the upper body, where the motion of the upper body is determined by a reference motion and assigned to the target motion. The model synthesizes the rest of the joints yielding a motion with the given sequence in the upper body, and with the learned motifs in the lower body.
A composition can be both spatial and temporal, and all it takes is an ROI mask where several frame sequences are zeroed, \ie, taken from the reference motion, and in the complementary part, several joints are zeroed (see Fig.~\ref{fig:composition_mask}).

\ifarxiv\else
\vspace{-5pt}
\fi

\subsection{Motion Harmonization}
\label{par:harmonization}
\ifarxiv
\begin{figure}%
\centering
\includegraphics[width=\linewidth]{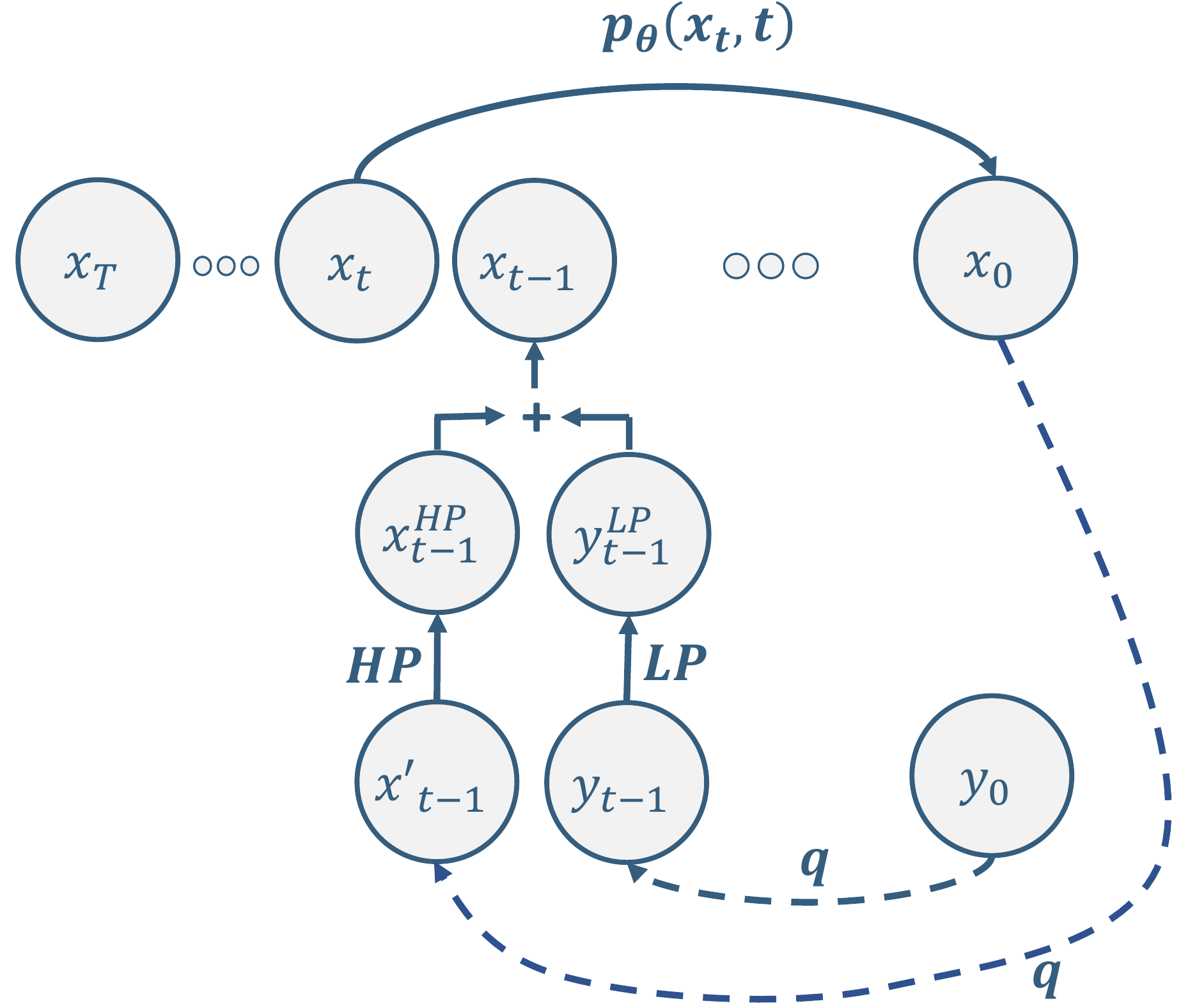} 

\vspace{-5pt}

\caption{\gt{Motion Harmonization. In order to inject guidance from the input motion $y_0$ during synthesis, we follow  \citet{choi2021ilvr} and add its low frequencies $y_{t-1}^{LP}$ at each denoising step $t$. }}
\label{fig:harmonization}
\end{figure}
\fi
Given a synthesized motion sequence $x_0$, we would like to integrate a portion of an unseen motion, $y$, into it. The portion of $y$ can be either temporal, \ie, several frames, or spatial, \ie, several joints, or both.
As visualized in Fig.~\ref{fig:harmonization}, \algoname{} overrides a window in $x_0$ with the desired portion of $y$ and denotes the outcome $y_0$.
Next, $y_0$ is harmonized such that it matches the core motion elements learned by our model, using a linear low-pass filter $\phi_N$ as suggested by \citet{choi2021ilvr}. 
\sr{Let $x'_{t-1}$ and $y_{t-1}$ denote the noised version of motions  $p_\theta(x_t,t)$ and  $y_0$, respectively.
The high-frequency details of $x'_{t-1}$ are added to the low-frequency of $y_{t-1}$ via
\begin{equation}
\ifarxiv\else \vspace{-5pt} \fi
    x_{t-1} = \phi_N(y_{t-1}) + x'_{t-1} - \phi_N(x'_{t-1}). 
\end{equation}
}

Note the difference between harmonization and motion composition: Both assign parts of an unseen sequence $y$ into a synthesized motion $x_0$. However, harmonization changes the assigned part such that it matches the learned distribution, while composition aims to keep it unchanged.

\ifarxiv\else
\vspace{-5pt}
\fi

\paragraph{Style transfer}
\ifarxiv
\ifarxiv 
\begin{figure}%
\else 
\begin{figure} 
\fi

\centering

\includegraphics[width=\columnwidth]{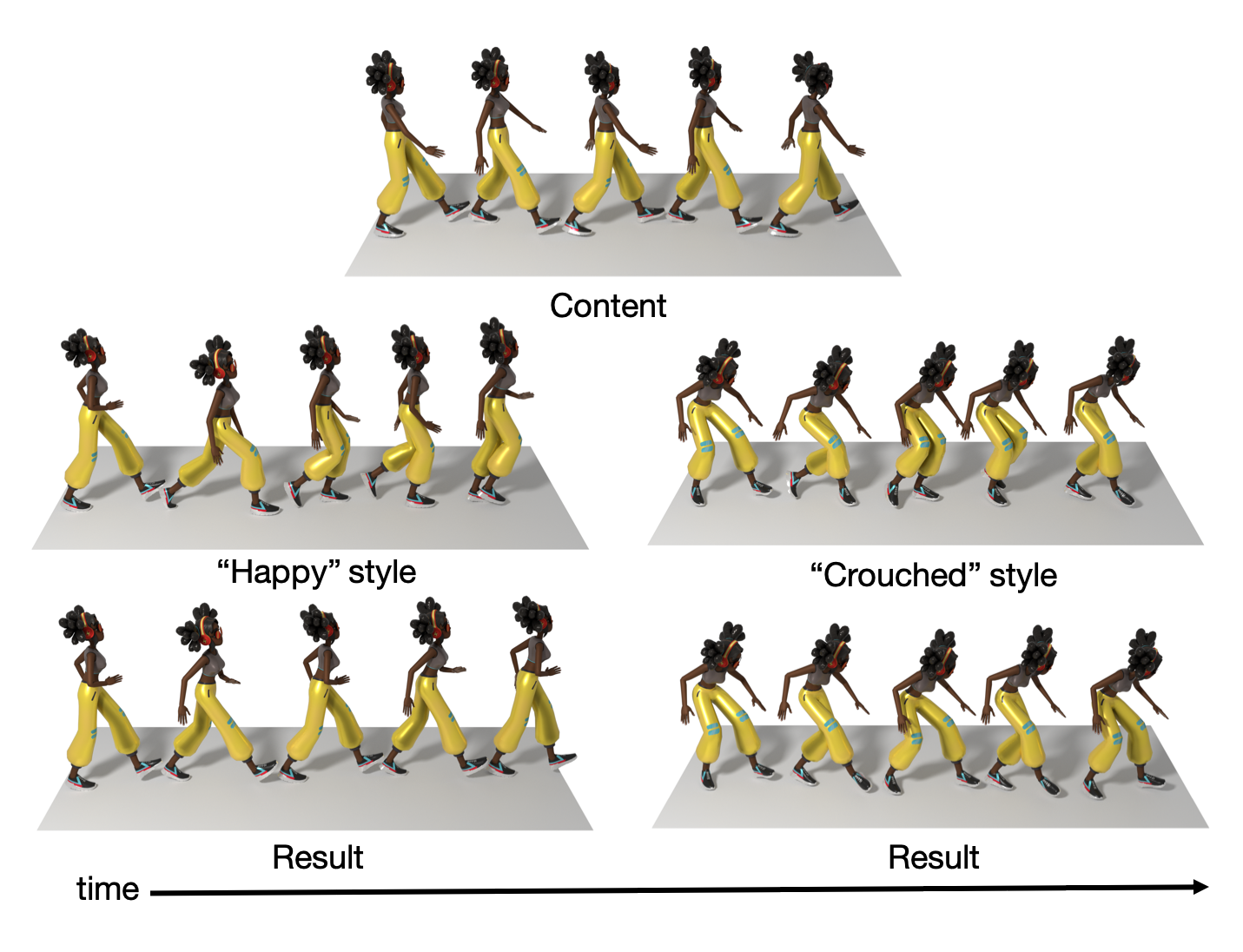}

\vspace{-10pt}
\caption{Style transfer is a special case of the harmonization application, where a reference motion is adjusted such that it matches the learned motion motifs. The style motion is learned by the model, and the content motion is unseen by it. Top: one content, unseen by the network, is applied to both styles. Left: a "happy" style, learned by the network, and below it the harmonized result. Right: a "crouched" style, learned by the network, and below it the harmonized result. Note that the character in both results is using the exact step rhythm and size as the character in the content motion.
}
\label{fig:style_tran}
\ifarxiv \end{figure}
\else \end{figure} \fi
\fi
We implement style transfer as a special case of harmonization, where instead of using a portion from $y$, we use all of it. That is, we fully override $x_0$.
We use a style motion $x$ learned by the model, and a content motion $y$, unseen by the model. Once applying harmonization, the result possesses the content of $y$ and the style of $x$, as depicted in Fig.~\ref{fig:style_tran}.

\ifarxiv\else
\vspace{-5pt}
\fi

\subsection{Straight-forward Applications}
\label{par:straightforward}
In this section we present applications that may require special techniques in existing works, but require no special technique when conducted using our model.
\srv{put refs}
\ifarxiv\else
\vspace{-5pt}
\fi

\paragraph{Long motion sequences}
\ifarxiv
\begin{figure}%
\centering
\includegraphics[width=\linewidth]{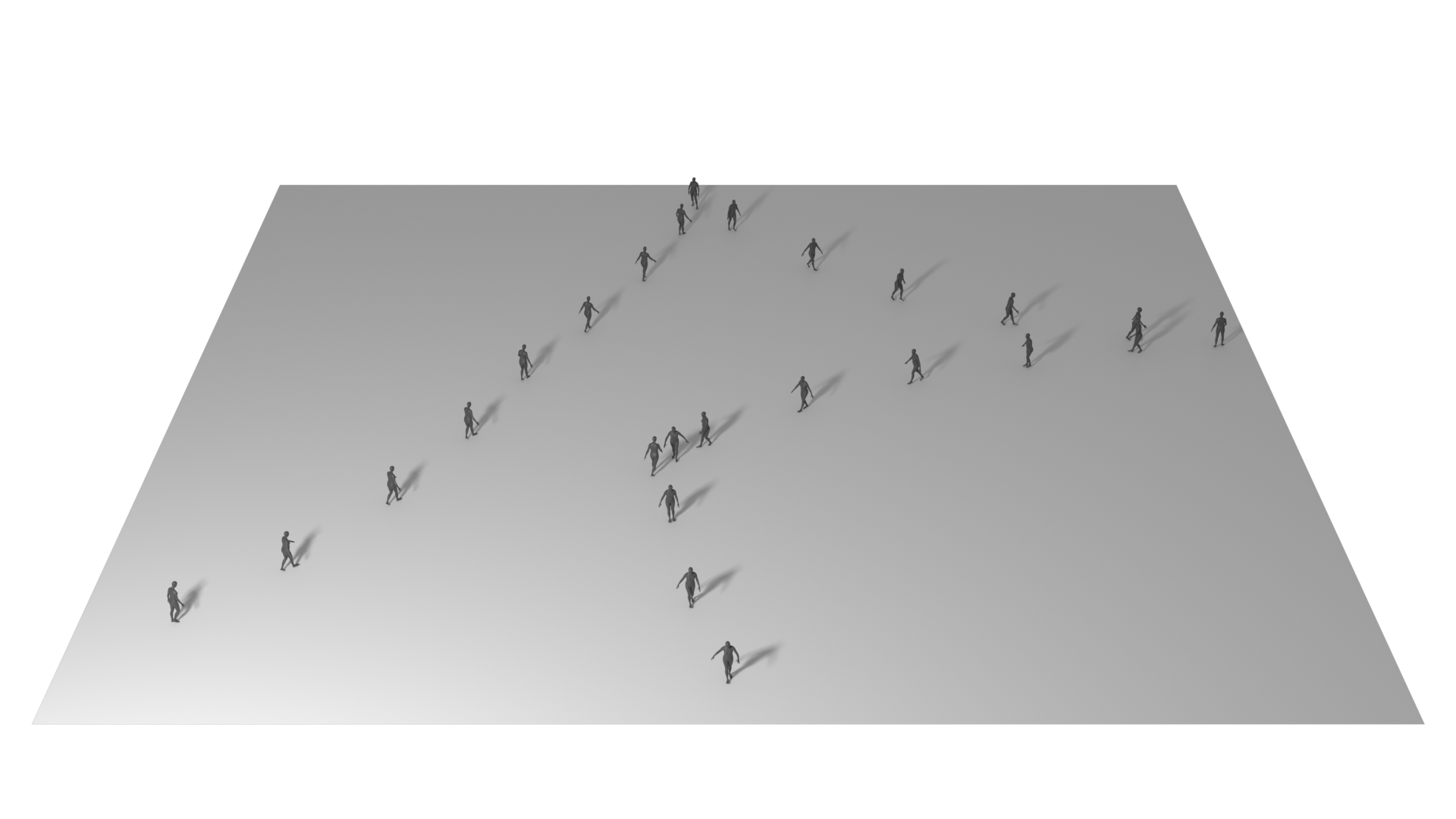}

\vspace{-10pt}
\caption{Long motion. The learned sequence is a 10 seconds motion, depicting a person walking back, then turning and walking back again. The synthesized motion is a 60 seconds sequence, depicting a person walking back, and occasionally turning and walking back again.}
\label{fig:long_motion}
\end{figure}
\fi
Our model can synthesize variable-length motions, even very long ones, with no additional training. Imputed to its small receptive field, the model can hallucinate a sequence as long as requested. An example of a one-minute animation is introduced in Fig.~\ref{fig:long_motion}. 

\ifarxiv\else
\vspace{-5pt}
\fi

\paragraph{Crowd animation}
\ifarxiv
\begin{figure}
\centering
\includegraphics[width=\linewidth]{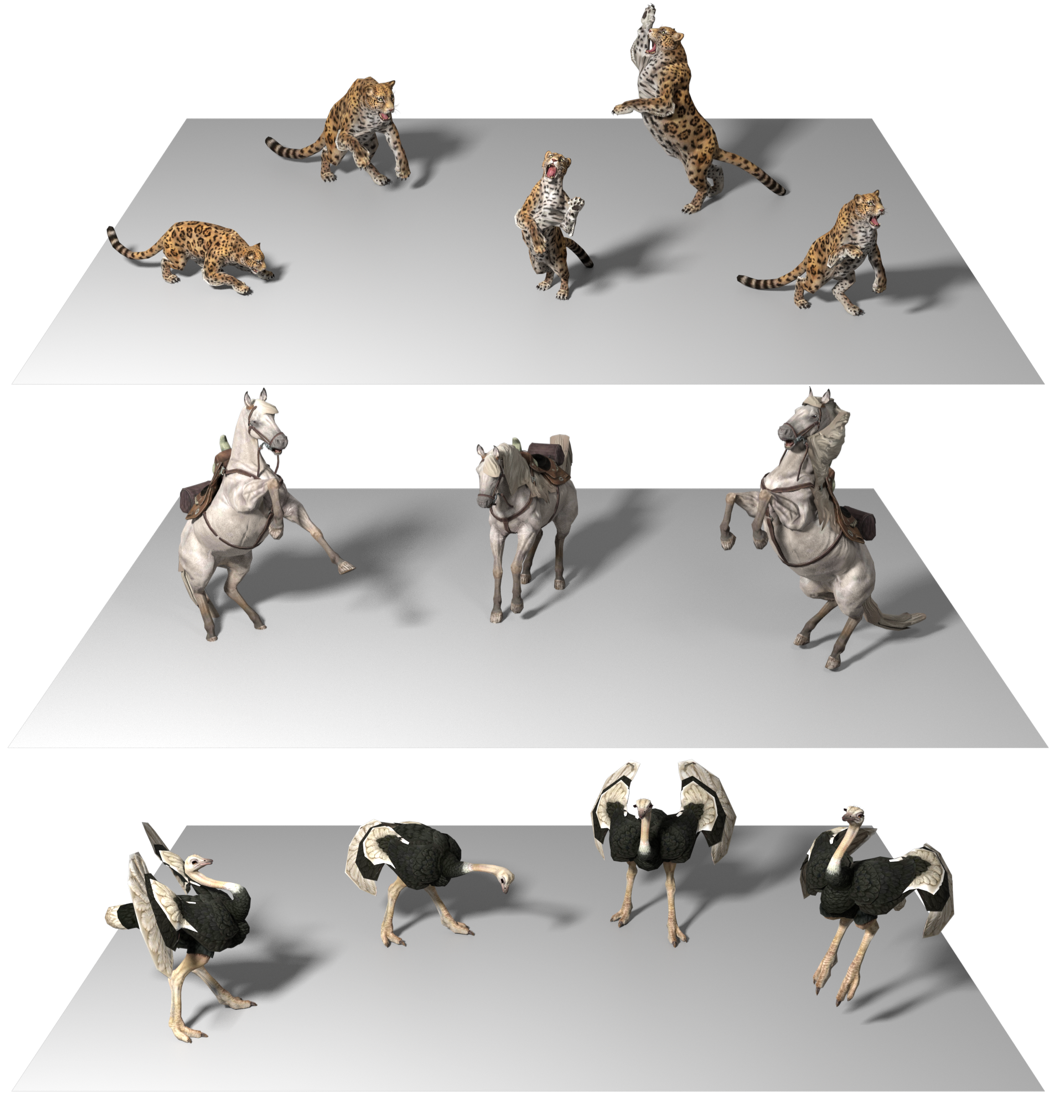}

\vspace{-10pt}
\caption{Crowd animation. Groups of jaguars, horses, and ostriches. In each group, no motion is like the other, and yet they are all learned from a single motion sequence.}
\label{fig:crowd}
\end{figure}
\fi
Although trained on a single sequence, during inference \algoname{} can generate a crowd performing a variety of similar motions, each sampled from a different Gaussian noise $x_{T} \sim \nn(0,I)$, as illustrated in Fig.~\ref{fig:crowd}.

\ifarxiv\else
\fi

\section{Experiments} \label{sec:experiments}

Our experiments are held on motion 
data from the HumanML3D~\shortcite{guo2022generating}, Mixamo~\shortcite{mixamo}, and Truebones Zoo~\shortcite{truebones} datasets, and on an artist-created animation, using an NVIDIA GeForce RTX 2080 Ti GPU.

\ifarxiv\else
\vspace{-5pt}
\fi

\subsection{Benchmarks}
We test our framework on two benchmarks. One consists of data from the HumanML3D dataset, and the other from the Mixamo dataset.
These two datasets are different in many aspects. 
The data in HumanML3D fits the SMPL \cite{loper2015smpl} topology, and its users normally use SMPL's mean body definition.
In contrast, Mixamo provides 70 characters, each possessing their unique bone lengths and some possessing unique topologies.  
In addition, the motions in the Mixamo dataset are more diverse and more dynamic. %

\ifarxiv\else
\vspace{-5pt}
\fi

\subsection{Metrics}
For each benchmark, we use a different set of metrics. For the Mixamo benchmark, we use the metrics introduced in Ganimator \cite{li2022ganimator}.
\sr{Ganimator is our immediate comparison reference, as it is the only single-motion learning work. For fairness, we compare with it using its own metrics. }
However, these metrics are based on the values of motion features (\eg, rotation angles) while the usage of deep features is the current best practice \cite{zhang2018unreasonable}. 
Given HumanML3D's capability for deep feature calculation, we utilize it to present our results specifically on these features.

\sr{
A \emph{good} score varies depending on the metric, being a \emph{high} value when higher is better and a \emph{low} value when lower is preferred.
Note that attaining a good score on some metrics, but a bad score on others, is insufficient: 
Good diversity scores with bad fidelity indicate deviation from the input motion, while good fidelity scores with bad diversity suggest overfitting.

An ideal outcome is a combination of good values for all metrics. 
For models with mixed scores, a better-scoring model is the one whose scores are more balanced. 
To this end, we follow established literature~\cite{rijsbergen1979information,chinchor1992muc} and suggest the Harmonic Mean metric, which is widely used in Machine Learning for this purpose \cite{taha2015metrics}. We compute it as follows: first, we normalize the scores for each metric. Normalization is between zero to the metric's maximum value. If the maximum is not known, we select the 90\% percentile of the computed scores. For metrics where lower is better, we subtract the score from the maximum value. Note that a negative value is therefore valid. We compute the Harmonic Mean via
\begin{equation}
    HM = E / \left(\frac{1}{s_1}+\dots+\frac{1}{s_E}\right),
\end{equation}

where $E$ is the number of metrics in a table and $s_i$ is the normalized score of metric $i$.
}

\ifarxiv\else
\vspace{-5pt}
\fi

\paragraph{Metrics on the Mixamo benchmark}

We use the Mixamo dataset to compare \algoname{} and Ganimator. 
For a fair comparison, we use the metrics suggested by them. However, their metrics do not measure the difference between synthesized motions (inter-diversity) nor the difference between sub-motions within one motion (intra-diversity), thus we add metrics to measure these missing qualities.
This group of metrics is applied to the motion itself, and not to deep features.

The metrics in Ganimator consist of (a) \emph{coverage}, which is the rate of temporal windows in the input motion $x_0$ that are reproduced by the synthesized one, (b) \emph{global diversity}, measuring the distance between $tess(\hat{x}_0$) and $x_0$, where $tess(\cdot)$ is a tessellation that minimizes the L2 distance to the input sequence, and (c) \emph{local diversity}, which is the average distance between windows in the synthesized motion $\hat{x}_0$ and their nearest neighbors in the input one.

The aforementioned metrics are measured relative to the input motion sequence. We add two metrics that are not related to the input motion, 
The first is (d) \emph{inter diversity}, the diversity between synthesized motions.
We define \emph{intra diversity} to be the diversity between sub-windows internal to a motion and define (e) \emph{intra diversity diff}, which is the difference between the intra diversity of the synthesized motions and  that of the input motion.

In addition, we measure time-space efficiency values: (f) the number of network parameters, (g) the number of required iterations, (h) the time required for each iteration, and (d) the total running time, which is a multiplication of the last two. 
For metrics (a)-(d), a higher score is better. 
For metrics (e)-(h), a lower score is better.

\ifarxiv\else
\vspace{-5pt}
\fi

\paragraph{Metrics on the HumanML3D benchmark}
We use this benchmark to measure metrics that are applied on deep features, obtained with a motion encoder by \citet{guo2022generating}.
The computed metrics are (a) \emph{SiFID} \cite{shaham2019singan}, which measures the distance between the distribution of sub-windows in the learned motion and a synthesized one, (b) \emph{inter diversity}, which is the LPIPS distance \cite{zhang2018unreasonable} between various motions synthesized out of one input, and (c) \emph{intra diversity diff}, which is the difference between the intra diversity of the synthesized motions and that of the input motion, where intra diversity is the LPIPS distance between sub-windows in one synthesized motion. 
For metrics (a) and (c) lower is better, and for metric (b) higher is better.

\ifarxiv\else
\vspace{-5pt}
\fi

\subsection{Quantitative Results}
\begin{table}
    \caption{Results on the Mixamo benchmark, comparing our work with state-of-the-art Ganimator. \algoname{} leads in all metrics but one. In particular, it demonstrates a significant advantage in the Harmonic Mean metric.
    }
    \vspace{-10pt}
    \centering
    \resizebox{\columnwidth}{!}{
    \begin{tabular} {l | c c c c c | c c c c | c }
        &  \rot{Coverage} $\uparrow$ & \rot{Global Div.} $\uparrow$ &  \rot{Local Div}. $\uparrow$ & \rot{Inter Div. } $\uparrow$& \rot{Intra Div. Diff.}  $\downarrow$  & \rot{\#Param. (M)} $\downarrow$ & \rot{\#Iter. (K)} $\downarrow$ &\rot{ Iter. Time (s)} $\downarrow$ & \rot{Tot. Time (h)} $\downarrow$ & \rot{Harmon. Mean} $\uparrow$ \\
        \midrule
        Ganimator & 94.3 & 1.24 & \textbf{1.17} & 0.09 & 0.13 & 21.7 & 60 (15$\times$4) & 0.36 & 6.0 & -0.22 \\
        SinMDM (Ours) & 94.3     
&   \textbf{1.42} &  1.00 & \textbf{0.13} & \textbf{0.03} & \textbf{5.26} & 60 & \textbf{0.09} & \textbf{1.5} & \textbf{0.85} \\
        \bottomrule
    \end{tabular}
    } %
    \label{tab:comp_mixamo_benchmark}
    \vspace{-5pt}
\end{table}

\begin{table}
    \caption{Results on the Gangnam-style motion. We mark the table leaders for the bottom part only, as MotionTexture and acRNN exhibit either overfit or divergence.
    In the lower part, where both models achieve high scores in all metrics, our model takes the lead, despite the fact that the subject motion has been selectively chosen by the authors of Ganimator.
    }
    \vspace{-10pt}
    \centering
    \resizebox{\columnwidth}{!}{
    \begin{tabular}{ l | c c c | c }
        &  Coverage $\uparrow$ & \makecell{Global \\ Diversity} $\uparrow$&  \makecell{Local \\ Diversity} $\uparrow$ & \makecell{Harmonic \\ Mean} $\uparrow$ \\
        \midrule
        MotionTexture~\shortcite{li2002motion}  &  84.6             & 1.03              & 1.04 & 0.32 \\
        MotionTexture (Single)                  & 100               & 0.21              & 0.33 & 0.09 \\
        acRNN~\shortcite{zhou2018auto}          & 11.6              & 5.63              & 6.69 & 0.30 \\
        \midrule
        Ganimator~\shortcite{li2022ganimator}   & 97.2              &  1.29             & \textbf{1.19} & 0.38 \\
   
        SinMDM (Ours) & \textbf{98.1} & \textbf{1.55} & 1.12 & \textbf{0.39} \\
        \bottomrule
    \end{tabular}
    } %
    \label{tab:quant_gangnam}
    \vspace{-5pt}
\end{table}

In table~\ref{tab:comp_mixamo_benchmark} we compare  
\algoname{} with Ganimator.
The table shows that our work outperforms Ganimator in all metrics except one. Specifically, \algoname{} exhibits a notable advantage in the Harmonic Mean metric, which effectively captures the collective strength of all scores.

All the metrics are computed separately on each benchmark motion and then averaged. The metrics that measure time were computed on benchmark motion number 9 only. 

The authors of Ganimator published a quantitative comparison  solely on one selected motion, namely the Gangnam-style dancing sequence.
We align with their study and measure our results on this motion as well, as shown in Tab.~\ref{tab:quant_gangnam}.
In this table we 
compare with two other, non-single motion works, MotionTexture \cite{li2002motion} and acRNN \cite{zhou2018auto}. 
Note that for this specific motion, selected by the Ganimator authors, our results lead the table in two metrics and are comparable in the third.

\ifarxiv
\begin{table}
    \caption{Comparison with motion diffusion model MDM. MDM achieves good SiFID and intra-diversity but exhibits poor inter-diversity, indicating overfitting. MDM on crops attains good inter-diversity but bad scores for the other metrics, indicating deviation from the input.
    Our model attains good scores in all metrics, demonstrating that a balance of good scores across all metrics is more important than excelling in only a select few.}
    \vspace{-10pt}
    \centering
    \resizebox{\columnwidth}{!}{
    \begin{tabular}{ l | c c c | c }
        &  SiFID $\downarrow$             & \makecell{Inter \\ Diversity} $\uparrow$         &        \makecell{Intra \\ Div. Diff.} $\downarrow$ & \makecell{Harmonic \\ Mean } $\uparrow$  \\
        \midrule
        MDM \shortcite{tevet2022human}  & 0.01  & 0.03 & 0.14 & 0.14 \\
        MDM on crops                    & 13.94 & 1.64 & 1.83 & -1.01 \\
        SinMDM (Ours)                   & 1.87  & 0.73 & 0.40 & \textbf{0.82} \\
        \bottomrule
    \end{tabular}
    } %
    \label{tab:quant_humanml}
    \vspace{-5pt}
\end{table}

To evaluate our performance vs. another motion diffusion model, we compare it with two variations of the MDM~\cite{tevet2022human} framework. 
The first is a vanilla MDM, trained on a single-motion.
The second is a variation of MDM in which we extract short crops out of the single-motion sequence and train an MDM on them. Note that the second variation holds a narrow receptive field.

The comparison is conducted on the HumanML3D dataset with metrics based on deep features. The results are shown in Tab.~\ref{tab:quant_humanml}. As mentioned above, attaining a high score in one metric only, indicates either overfit or divergence from the input motion. MDM yields complete overfit, thus its SiFID and intra-diversity scores are perfect (indicating similarity to the input motion), but its inter-diversity scores are low.
The overfit of MDM is caused by the global attention it uses.
On the other hand, the quantitative results for the second MDM variation indicate divergence from the input motion motifs.
These quantitative results are supported by the qualitative results in our supplementary video.
\else

\fi

\begin{figure}%
\centering
\includegraphics[width=.8\linewidth]{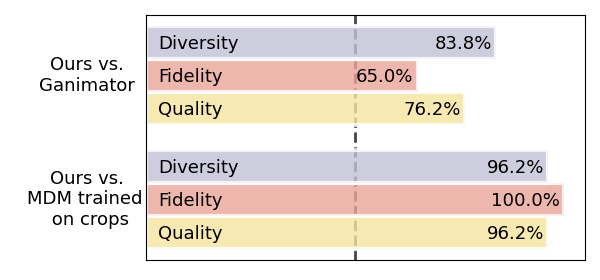}

\vspace{-10pt}

\caption{User study. Users vote that our model performs better than state-of-the-art Ganimator and MDM trained on crops.
The dashed line marks 50\%.}
\label{fig:user_study}
\vspace{-5pt}
\end{figure}
\sr{
Finally, 
we perform a user study in which users are requested to judge which model is better in terms of diversity, fidelity, and quality. 
In the study, we compare our model vs. Ganimator and MDM trained on crops.
Each pair of models is compared over 8 different motions, and each such comparison is judged by 10 distinct users.
The results (Fig. \ref{fig:user_study}) show that our model is significantly preferred by the users.
Screenshots from our user study are provided in \ifappendix{Appendix~\ref{app:ustudy}}\else{the sup. mat}\fi.
}

\ifarxiv\else
\vspace{-5pt}
\fi

\subsection{Qualitative results}
\ifarxiv
\begin{figure}%
\centering
\includegraphics[width=\linewidth]{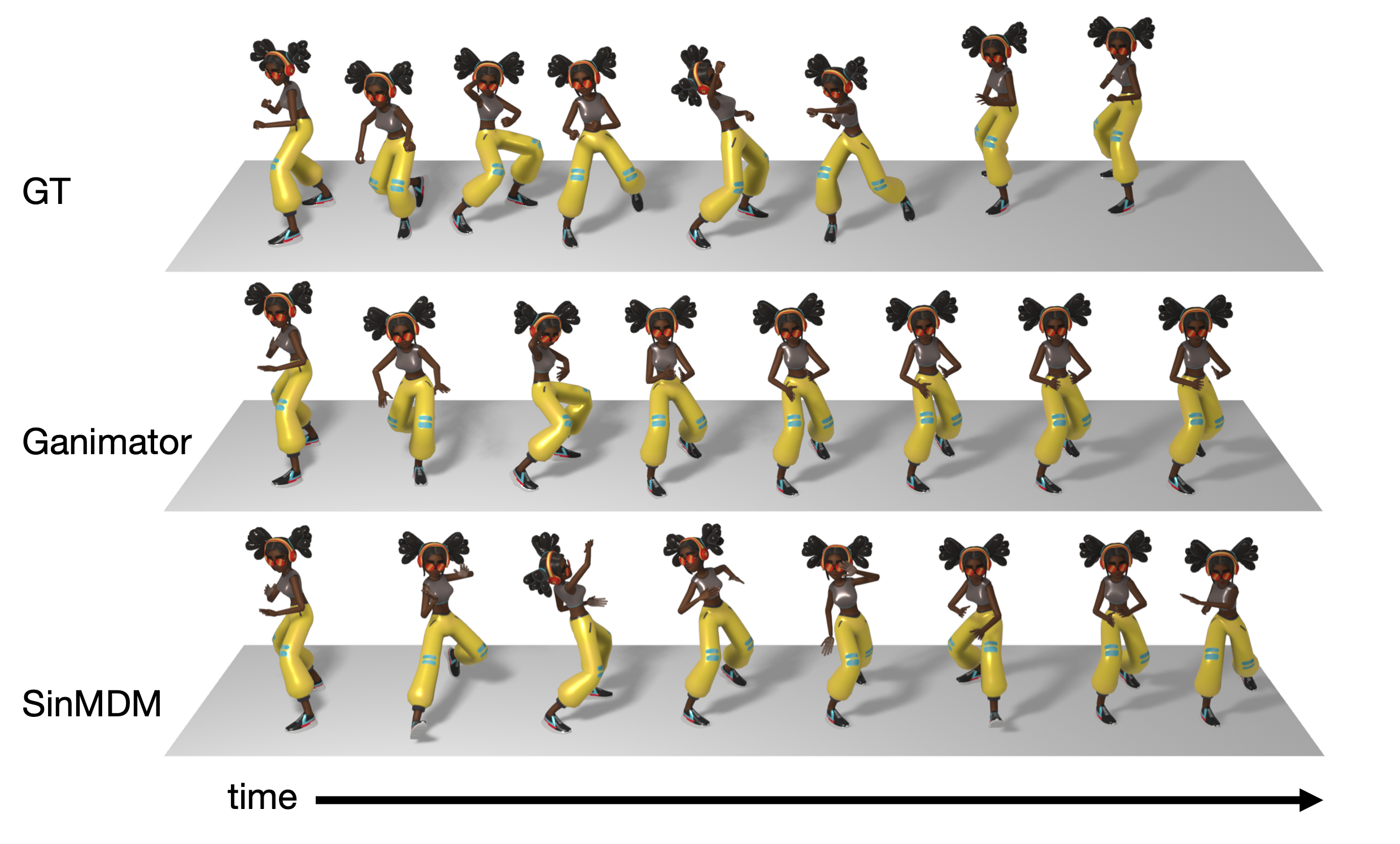} 

\vspace{-10pt}

\caption{Qualitative comparison, based on the motion \emph{punch to elbow} from the Mixamo dataset. Observe the mode collapse in Ganimator's synthesized motion, where over half of the motion is frozen.}
\label{fig:qualit_compare}
\end{figure}

\begin{figure}%
\centering
\includegraphics[width=\linewidth]{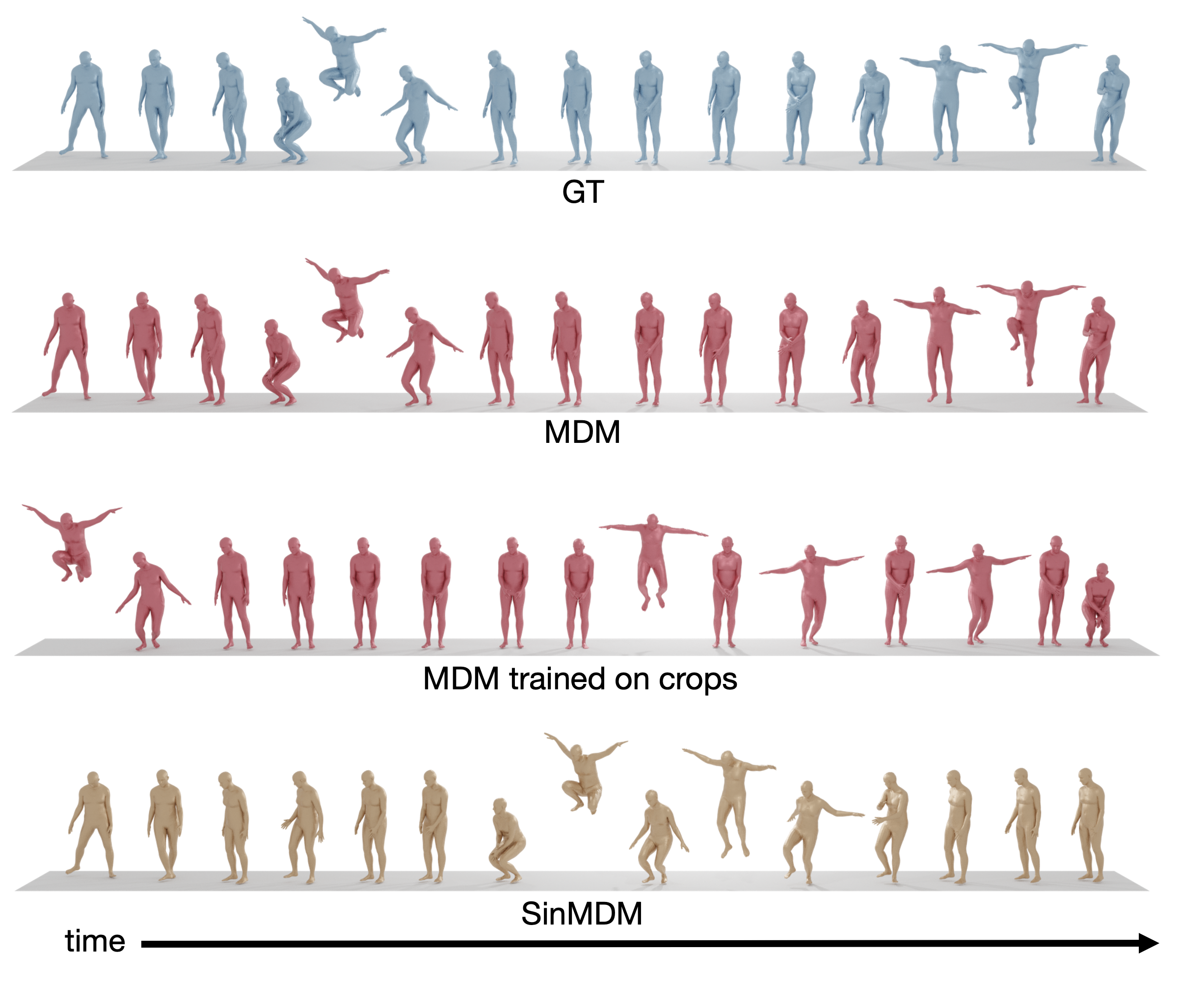} 

\vspace{-10pt}

\caption{Qualitative comparison on the HumanML3D dataset. 
MDM exhibits overfit, and MDM trained on crops exhibits jittery motion, \eg, when transferring from standing to jumping without bending the knees before and after.
}
\label{fig:qualit_compare_mdm}
\end{figure}
\fi
Our supplementary video reflects the quality of our results. 
It presents multiple synthesized motions, as well as a comparison to other works.
In addition, Fig.~\ref{fig:qualit_compare} and \ref{fig:qualit_compare_mdm} depict SinMDM vs. current work. 
In contrast to other works that exhibit mode collapse, overfitting, or produce jittery motion, SinMDM demonstrates none of these issues.

\ifarxiv\else
\vspace{-5pt}
\fi

\subsection{Ablation} \label{sec:ablation}
\begin{table}
    \caption{Ablation results on the HumanML3D benchmarks. Our selected architecture is framed in gray.
    Rows 1,2: comparing receptive field widths. 
    Rows 3,4:  vanilla attention vs. none (vs. QnA in row 2). 
    Row 5: QnA-based transformer (vs. QnA-based UNet in row 2). 
    Row 1 indicates overfit 
    and row 6 indicates divergence. 
    Rows 3,4 present good scores but not as good as ours.
    Abbr.: \emph{r.f.} $\rightarrow$ receptive field, \emph{d} $\rightarrow$ depth, \emph{atn.} $\rightarrow$ attention.
    }
    \vspace{-10pt}
    \centering
    \resizebox{\columnwidth}{!}{
    \begin{tabular}{ l | c c c | c}
        &  SiFID $\downarrow$ & \makecell{Inter  \\ Diversity} $\uparrow$ & 
       \makecell{Intra \\ Div. Dist.} $\downarrow$ & \makecell{Harmonic \\ Mean} $\uparrow$ \\
        \midrule
        UNet w/ QnA & &  & \\
        \quad wide r.f. (d=3)  &  0.69 &  0.20  & 0.34 & 0.54 \\
        \arrayrulecolor{gray}\hline
        \multicolumn{1}{|c|}{\quad \kern-1em narrow r.f. (d=1)} & 1.87 & 0.73 & 0.40 & \multicolumn{1}{c|}{\textbf{0.82}} \\
        \hline \arrayrulecolor{black}
        UNet (d=1) & & & \\
        \quad w/ vanilla atn. & 1.88 & 0.72 &  0.45 & 0.79 \\
        \quad w/o atn. w/o QnA & 2.03 & 0.72 & 0.43 & 0.79\\
        Transformer w/ QnA & 5.99 &  1.74 & 0.57 & 0.56\\
        \bottomrule
    \end{tabular}
    } %
    \label{tab:ablation}
\end{table}

We examine several architectural variations on the HumanML3D benchmark and present the results in Tab.~\ref{tab:ablation}.
We start by confirming that a narrow receptive field produces plausible results while a wider one induces overfit (rows 1,2). In order to do so, we examine a fixed architecture (QnA-based UNet) with two receptive field widths. We control the width by tweaking the depth of the UNet.
Indeed we observe that the model with the wide receptive field overfits (replicates) the input motion, as its inter-diversity is bad while its SinFid and intra-diversity are good.

Recall that the UNet architecture used by many diffusion models~\cite{nichol2021glide,ho2020denoising} holds global attention layers in it.
In the next experiment (rows 3,4), we confirm that replacing UNet's global attention with a local one (QnA) is a good choice. We fix the network's depth and examine alternatives to QnA.
One alternative is the usage of vanilla attention, and the other is to use no attention whatsoever.
Both alternatives show plausible metric results, and yet, our QnA-based UNet (row 2) performs better.
Plausible results with vanilla attention (row 3) are noteworthy, considering its global receptive field. 
This can be attributed to the absence of temporal positional embedding,
enabling the generative model to identify motion patterns across various temporal regions.

Finally, as many motion diffusion models favor a transformer over a UNet~\cite{tevet2022human,kim2022flame}, we measure the scores for a QnA-based transformer (row 5). 
To refrain from overfitting, we apply QnA layers within the transformer as we do with the UNet.
In addition, to promote diversity and permit the rearrangement of motion components, we employ relative temporal positional embeddings~\cite{shaw2018self,press2021train,su2021roformer} instead of the existing global ones.
However, the QnA-based transformer attains a bad SiFiD score, indicating poor fidelity to the input motion.

Note that due to the mixed scores (that indicate either overfit or divergence), the usage of the Harmonic Mean metric is essential as it allows for the assessment of the combined strength of all scores.

\ifarxiv\else
\vspace{-5pt}
\fi

\section{Conclusions} \label{sec:conclusion}

We have explored the use of diffusion models on single motion sequence synthesis and designed a motion denoising transformer with a narrow receptive field. 
Training on single motions is particularly useful in motion domains, where the number of data instances is scarce. Particularly, for animals and imaginary creatures, which have unique skeletons and motion motifs. The motion of such creatures cannot be captured easily nor learned from the human motion data available.  

Our experiments on several datasets demonstrate that our lightweight diffusion-based method significantly outperforms current work both in quality and time-space performance.
Moreover, our approach allows the synthesis of particularly long motions, and enables a variety of motion manipulation tasks, including spatial and temporal in-betweening, motion expansion, harmonization, style transfer, and crowd animation. 

The innate limitation of our method,  common to all models (in all domains) that learn a single instance, is the limited ability to synthesize out-of-distribution.
However, the main limitation of our diffusion-based approach is the relatively long inference time. This is due to the iterative nature of diffusion models. 

Finally, our work shows the competence of diffusion models to learn from limited data, which contradicts their reputation for requiring large amounts of data.
Nevertheless, in the future, we would like to address the single input limitations, by possibly learning from available motion data of creatures with rather compatible skeletons.

\ifarxiv
    \section{Acknowledgments}
We are grateful to 
Panayiotis Charalambous, 
Andreas Aristidou and Brian Gordon for reviewing earlier versions of the manuscript.
This research was supported in part by the Israel Science Foundation (grants no. 2492/20 and 3441/21), Len Blavatnik and the Blavatnik family foundation, and the Tel Aviv University Innovation Laboratories (TILabs).

\fi

\bibliographystyle{ACM-Reference-Format}
\bibliography{main}

\ifarxiv
\else
    \clearpage
    \begin{figure}%
\centering
\includegraphics[width=\linewidth]{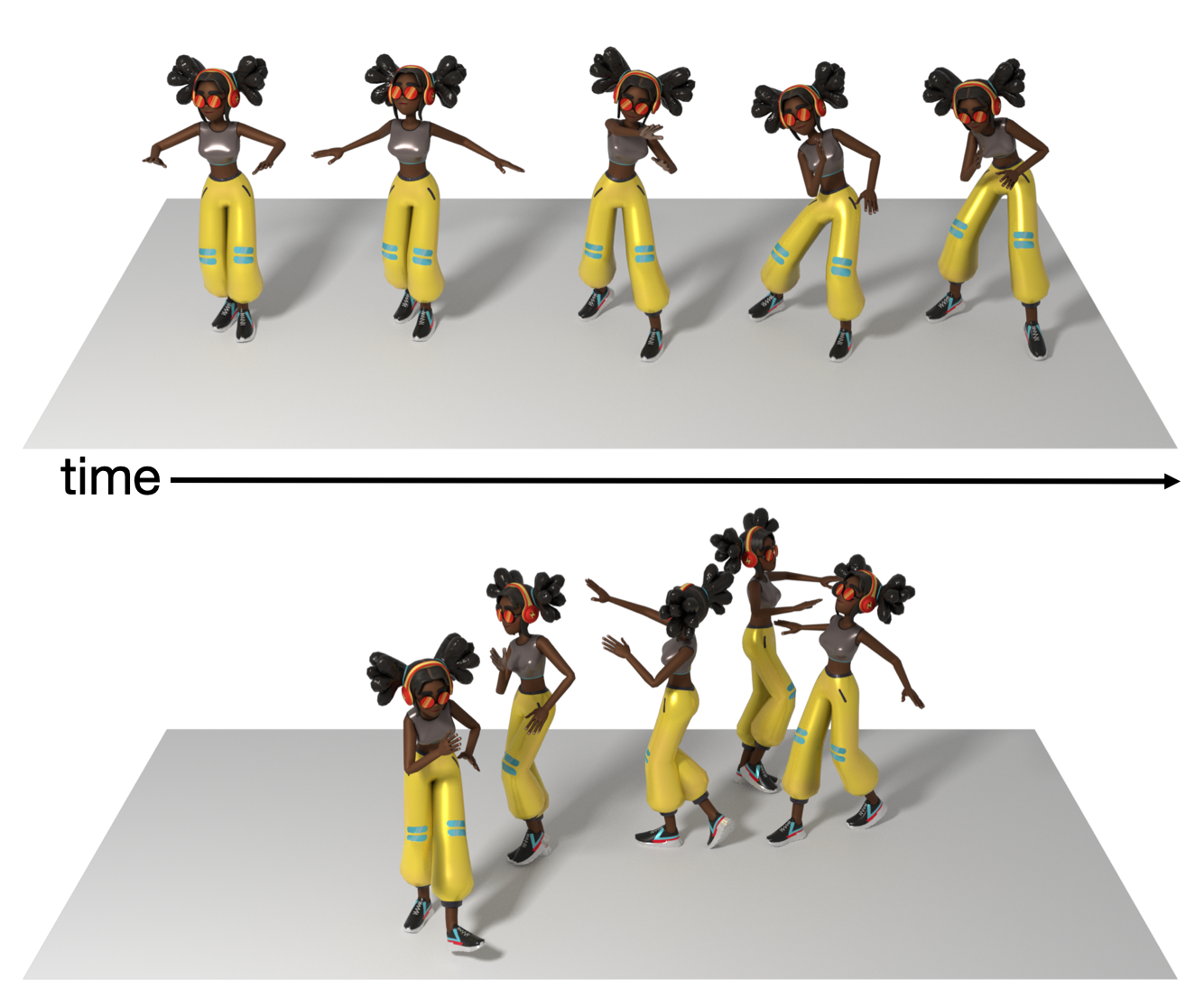} 

\vspace{-10pt}

\caption{Spatial composition. Top: reference motion, unseen by the network. Bottom: composed motion. 
The referenced sequence is \emph{warm-up}, and the learned one is \emph{walk in circle}. In the composed result, the top body part performs a warm-up activity, and the bottom body part walks in a curved path.
}
\label{fig:spatial_comp}
\end{figure}
    \begin{figure}
\centering
\includegraphics[width=\linewidth]{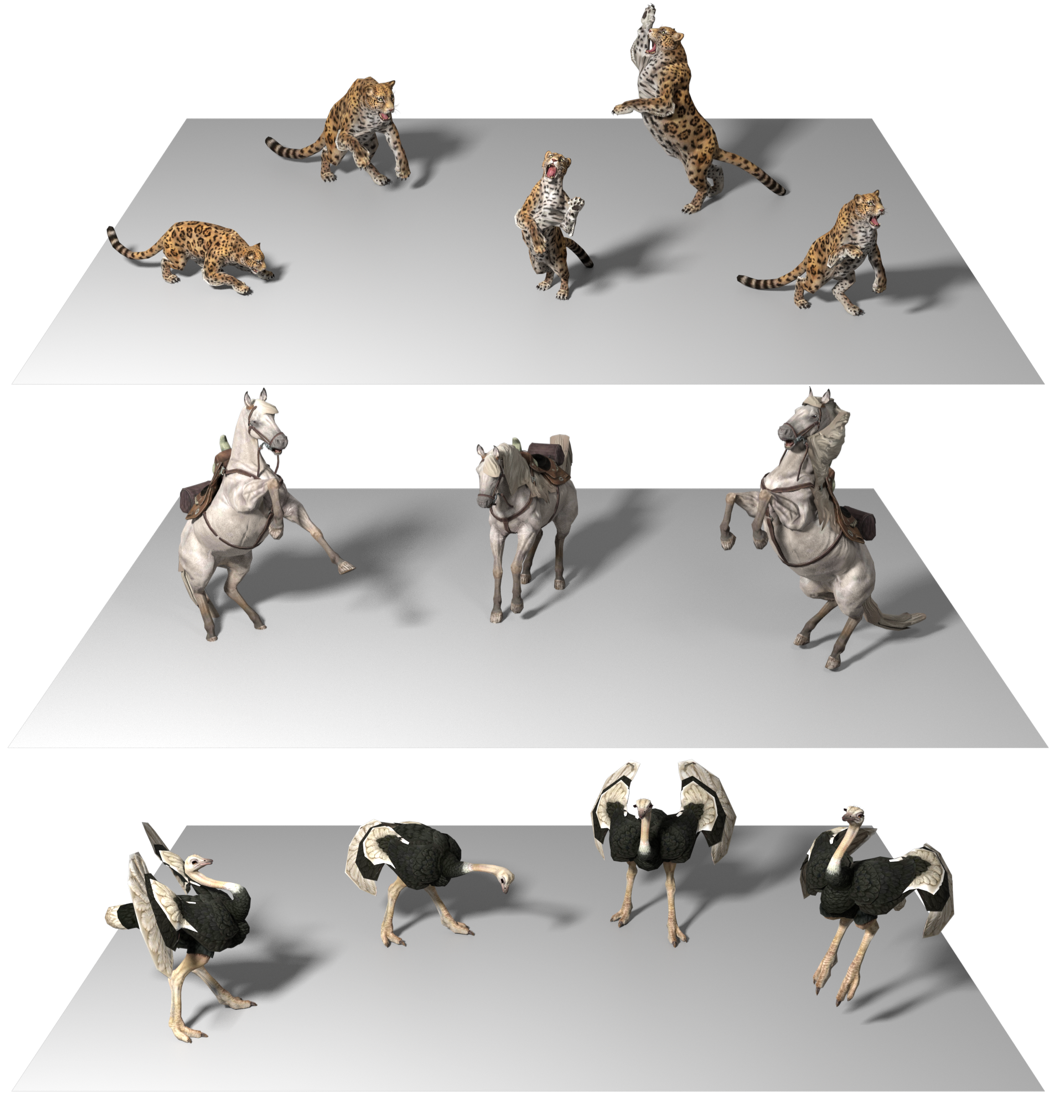}

\vspace{-10pt}
\caption{Crowd animation. Groups of jaguars, horses, and ostriches. In each group, no motion is like the other, and yet they are all learned from a single motion sequence.}
\label{fig:crowd}
\end{figure}
    \ifarxiv 
\begin{figure}
\else 
\begin{figure} 
\fi

\centering

\includegraphics[width=\columnwidth]{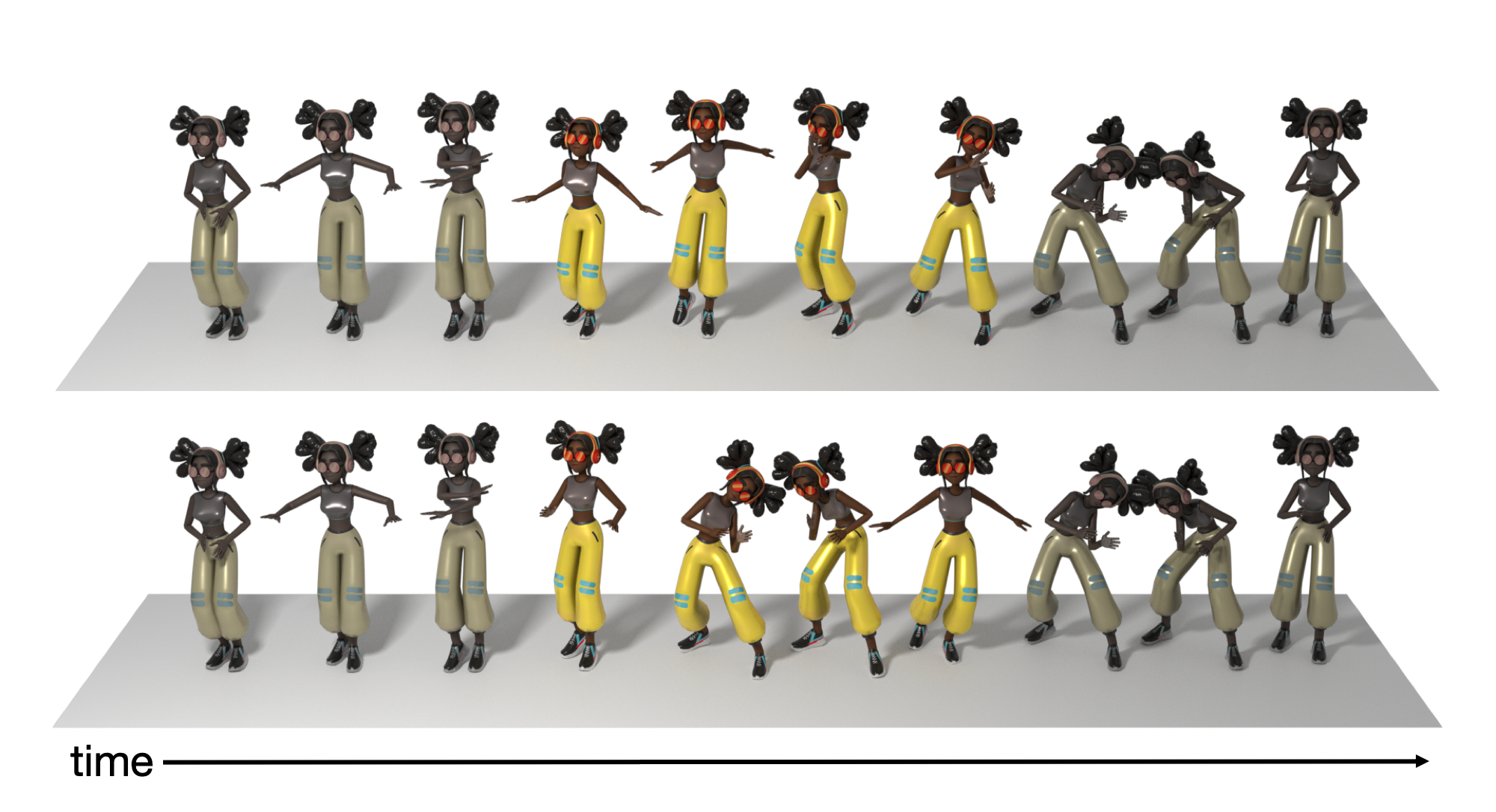}

\vspace{-10pt}
\caption{Temporal composition -- In-betweening. Both top and bottom show results for the same input, introducing diverse outputs. The beginning and the end of the motion are given by the reference sequence and can be distinguished according to their faded tone. Observe that the beginning and the end are identical in both sequences. The center of each motion is synthesized.  }
\label{fig:temporal_comp_inbetween}

\ifarxiv 
\end{figure}
\else 
\end{figure} 
\fi

    \begin{figure}%
\centering
\includegraphics[width=\linewidth]{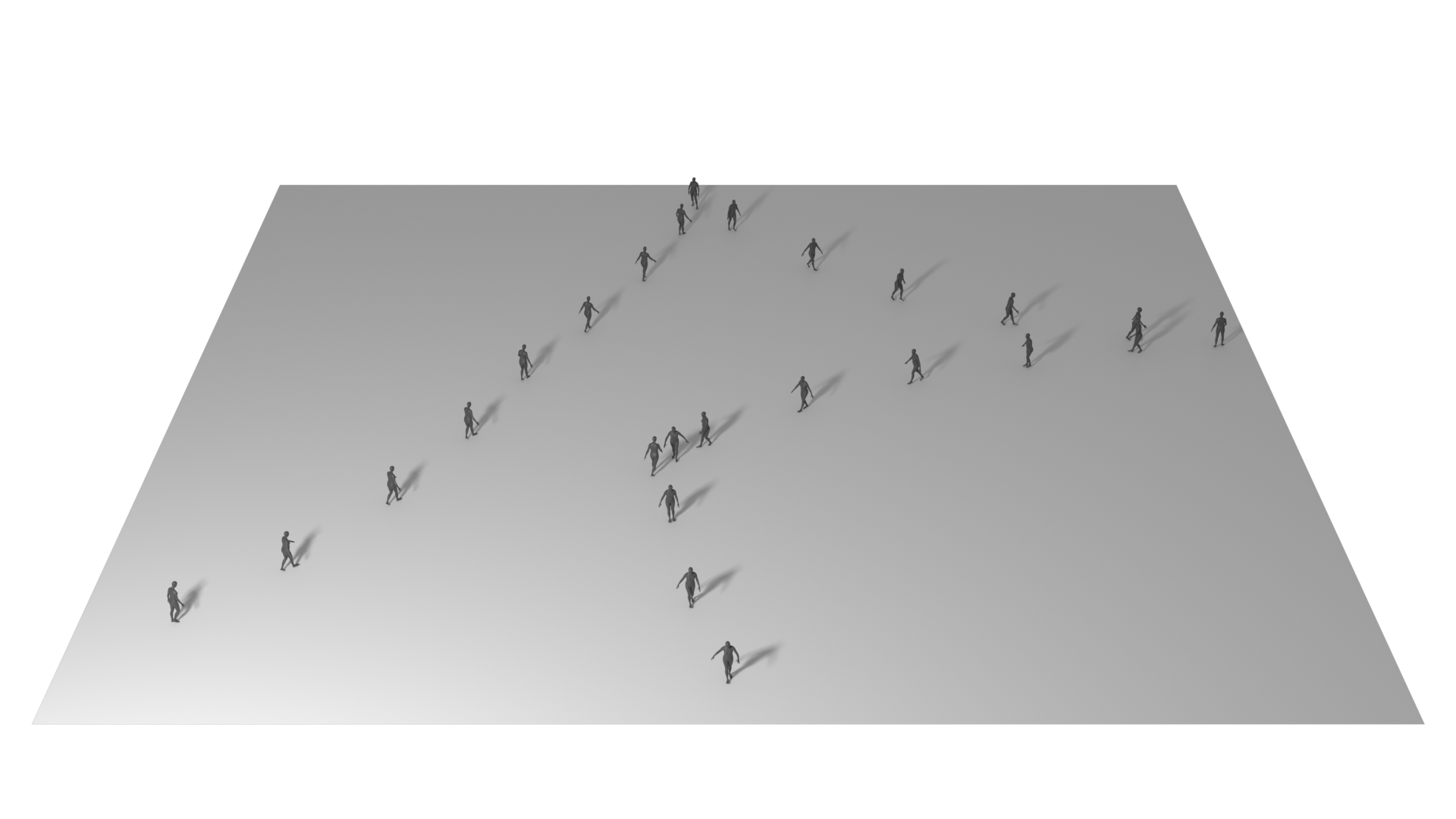}

\vspace{-10pt}
\caption{Long motion. The learned sequence is a 10 seconds motion, depicting a person walking back, then turning and walking back again. The synthesized motion is a 60 seconds sequence, depicting a person walking back, and occasionally turning and walking back again.}
\label{fig:long_motion}
\end{figure}
    \begin{figure}%
\centering
\includegraphics[width=\linewidth]{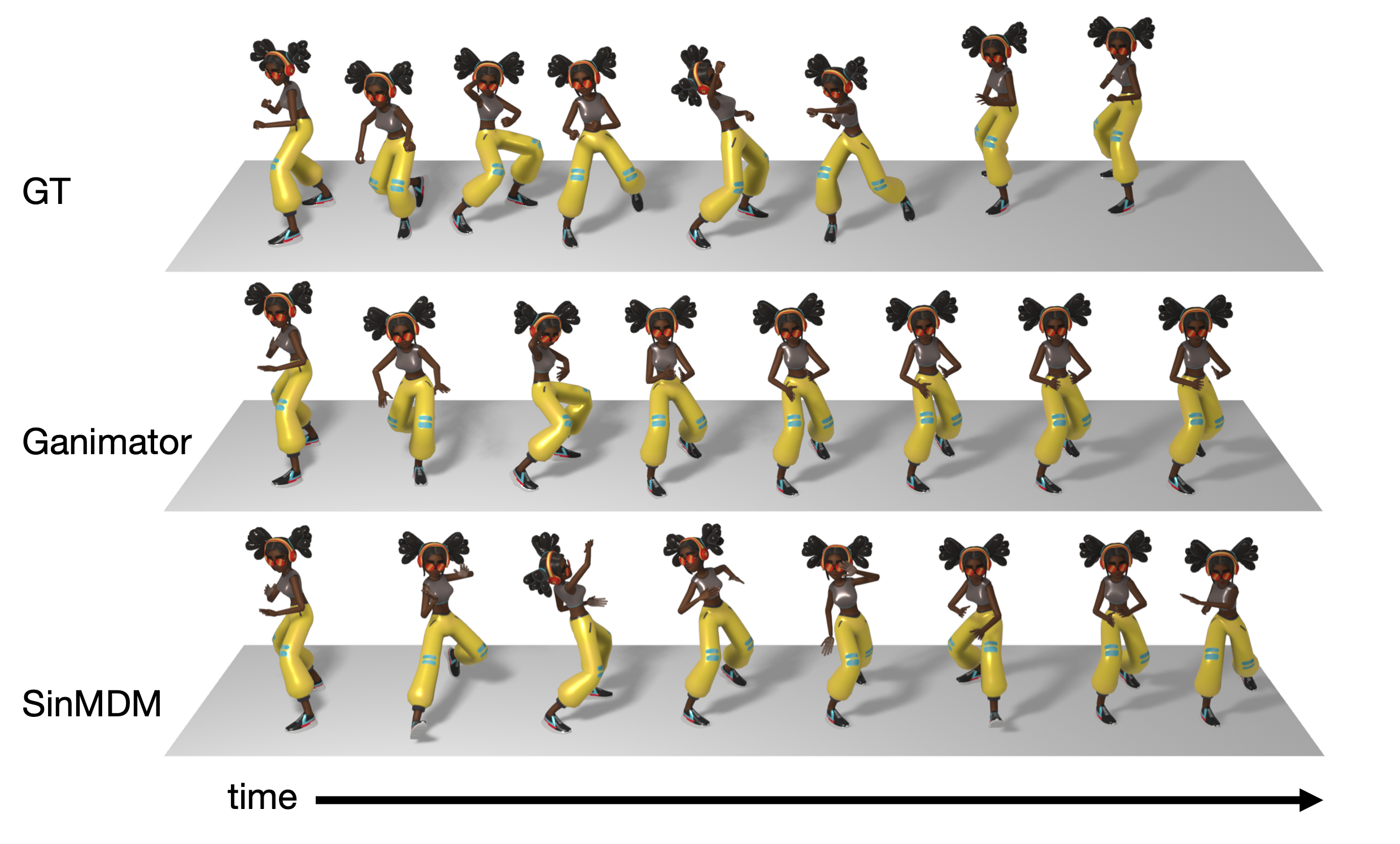} 

\vspace{-10pt}

\caption{Qualitative comparison, based on the motion \emph{punch to elbow} from the Mixamo dataset. Observe the mode collapse in Ganimator's synthesized motion, where over half of the motion is frozen.}
\label{fig:qualit_compare}
\end{figure}

    \ifarxiv 
\begin{figure}%
\else 
\begin{figure} 
\fi

\centering

\includegraphics[width=\columnwidth]{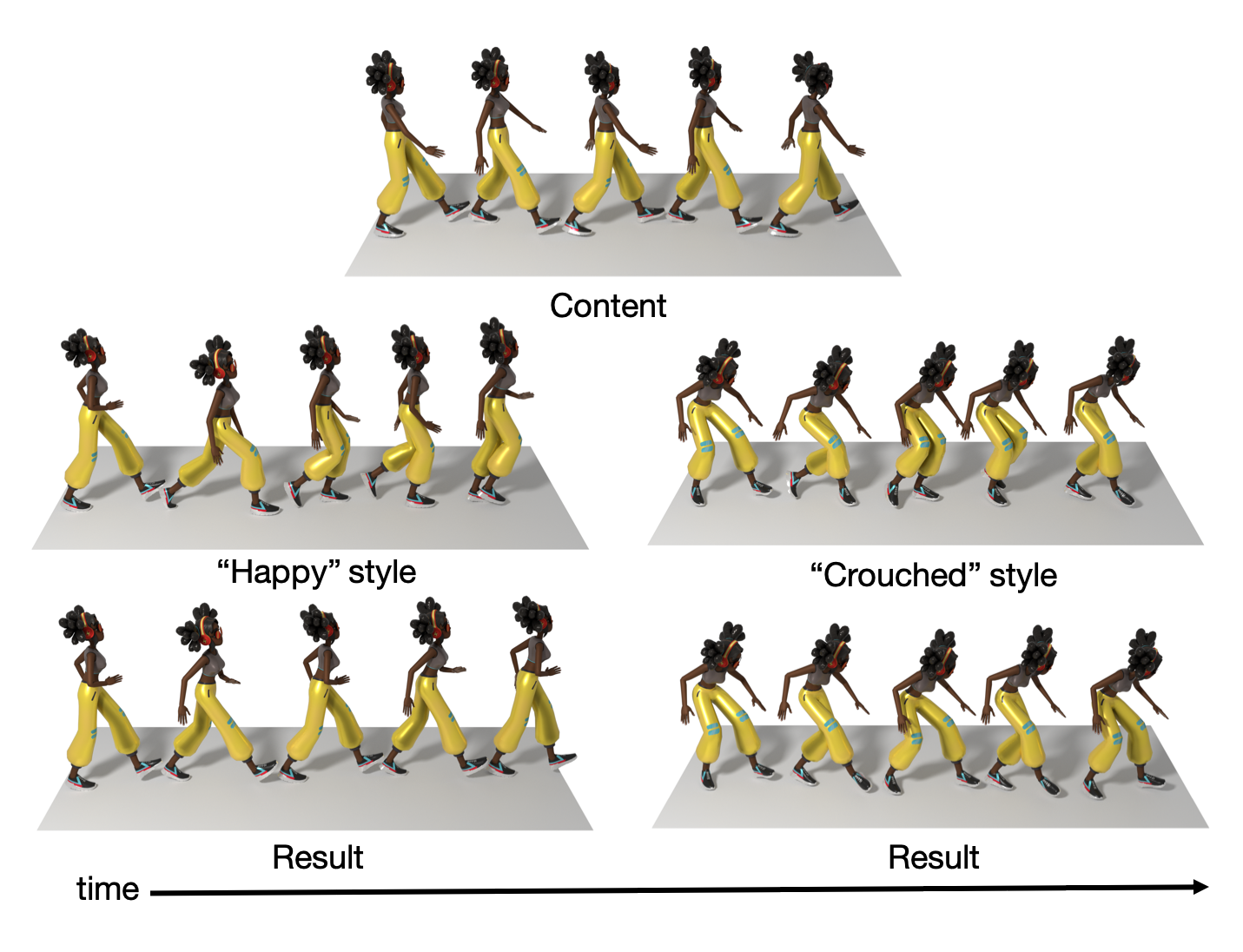}

\vspace{-10pt}
\caption{Style transfer is a special case of the harmonization application, where a reference motion is adjusted such that it matches the learned motion motifs. The style motion is learned by the model, and the content motion is unseen by it. Top: one content, unseen by the network, is applied to both styles. Left: a "happy" style, learned by the network, and below it the harmonized result. Right: a "crouched" style, learned by the network, and below it the harmonized result. Note that the character in both results is using the exact step rhythm and size as the character in the content motion.
}
\label{fig:style_tran}
\ifarxiv \end{figure}
\else \end{figure} \fi
    \begin{figure}%
\centering
\includegraphics[width=\linewidth]{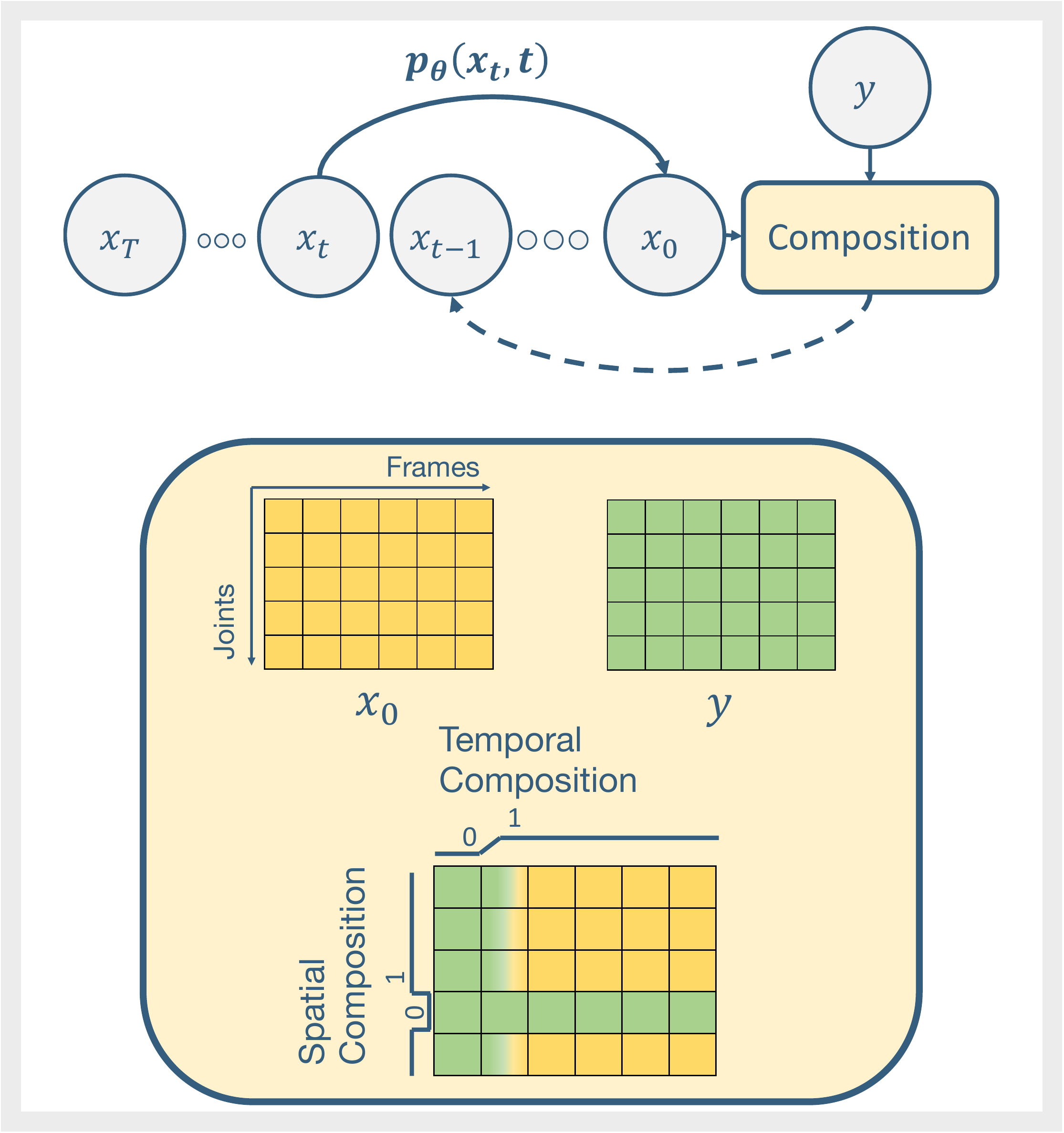} 

\vspace{-5pt}

\caption{Motion composition. Parts from a reference motion $y$, are composed with the synthesized motion $\hat{x}_0$, according to a composition map. }
\label{fig:composition_mask}
\end{figure}
    
    \clearpage
\fi

\ifappendix
    \vspace{50pt}
    \centerline{\LARGE\bfseries Appendix\par}
    \appendix
    \begin{figure*}[t]
    \centering
    \begin{subfigure}[c]{0.48\textwidth}
        \includegraphics[height=2.1cm]{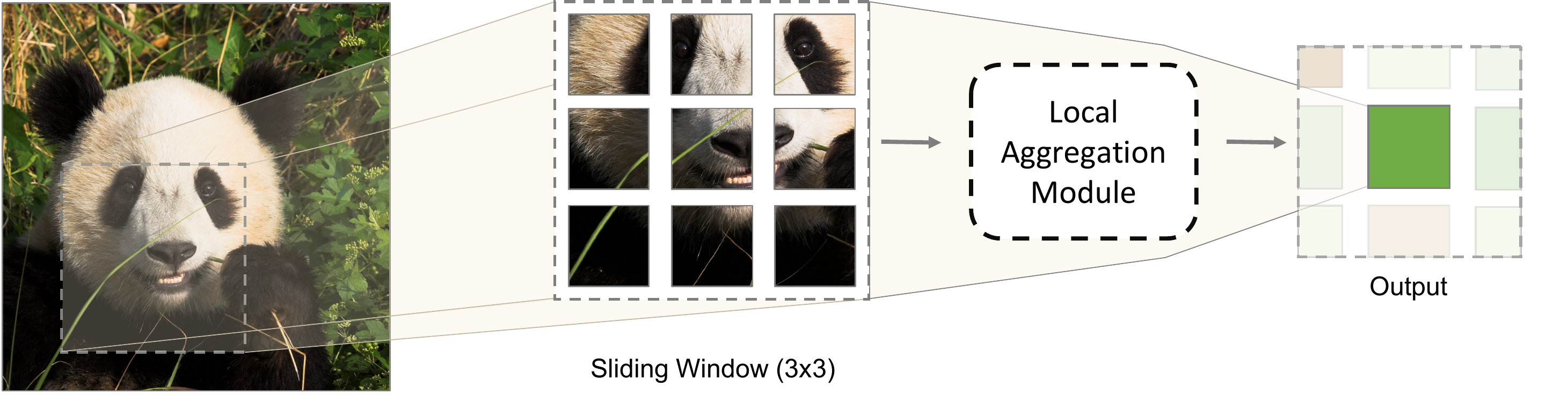}
    \end{subfigure}
     \begin{subfigure}[c]{0.15\textwidth}
        \includegraphics[height=3.75cm]{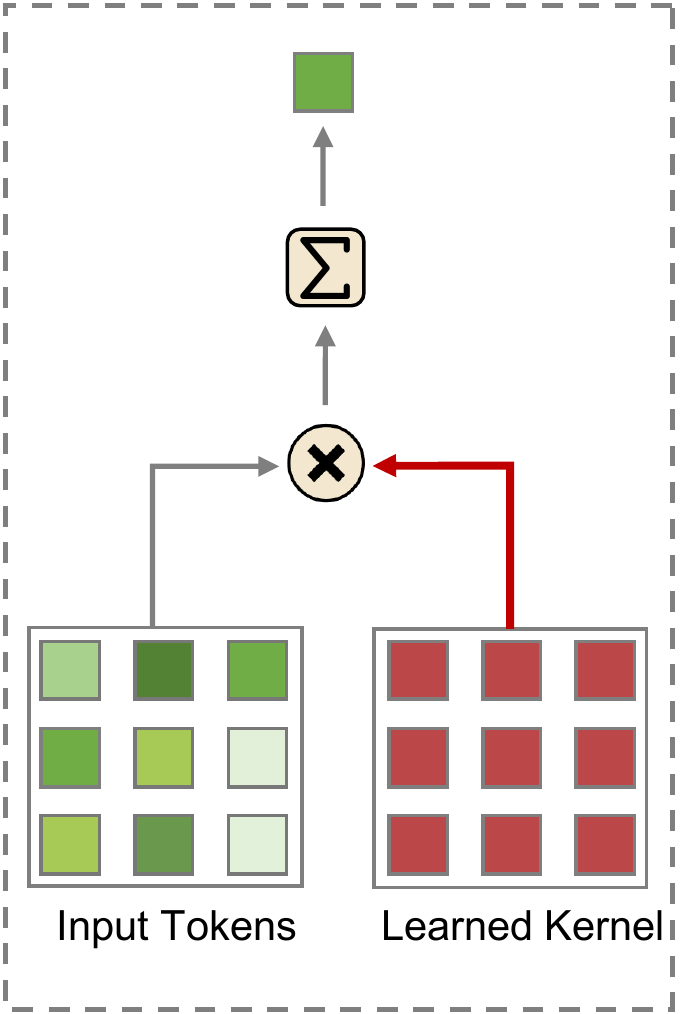}
        \vspace{-5pt}
        \subcaption{Convolution}
        \label{subfig:qna_overview_conv}
    \end{subfigure}
     \begin{subfigure}[c]{0.15\textwidth}
        \includegraphics[height=3.75cm]{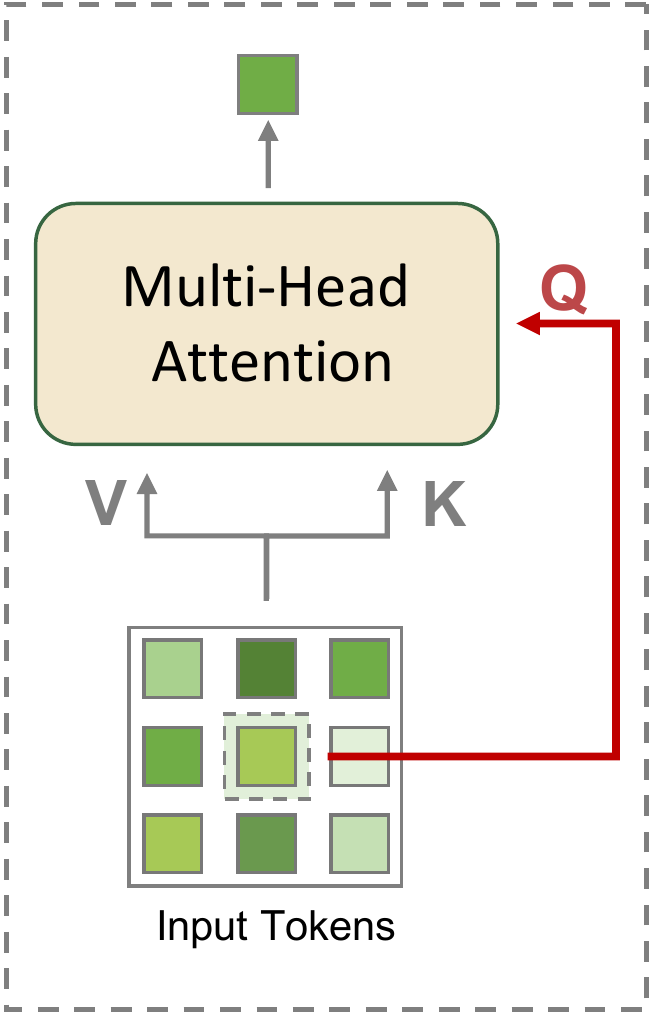}
        \vspace{-5pt}
        \subcaption{SASA~\shortcite{SASA}}
        \label{subfig:qna_overview_sasa}
    \end{subfigure}
     \begin{subfigure}[c]{0.17\textwidth}
        \includegraphics[height=3.75cm]{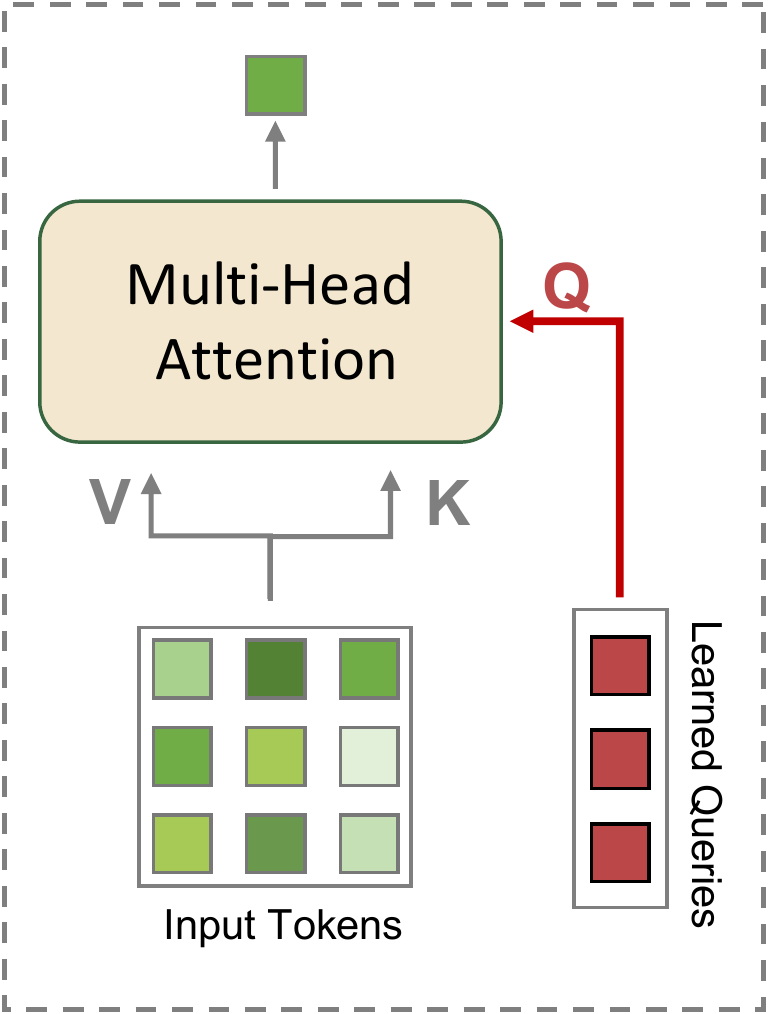}
        \vspace{-5pt}
        \subcaption{QnA~\shortcite{arar2022learned}}
        \label{subfig:qna_overview_ours}
    \end{subfigure}
    \vspace{-5pt}
    \caption{
    QnA overview (extracted from the QnA paper).
    Left: Local layers may utilize various approaches to overlapping windows.
    (a) Convolutions apply aggregation by learning shared weighted filters. 
    (b) SASA~\shortcite{SASA} combines window tokens through self-attention. 
    (c) QnA use shared learned queries across windows, maintaining the expressive power of attention while achieving linear space complexity.}
    \label{fig:qna_overview}
\end{figure*}
\begin{figure}[t]
    \centering
    \includegraphics[width=\linewidth]{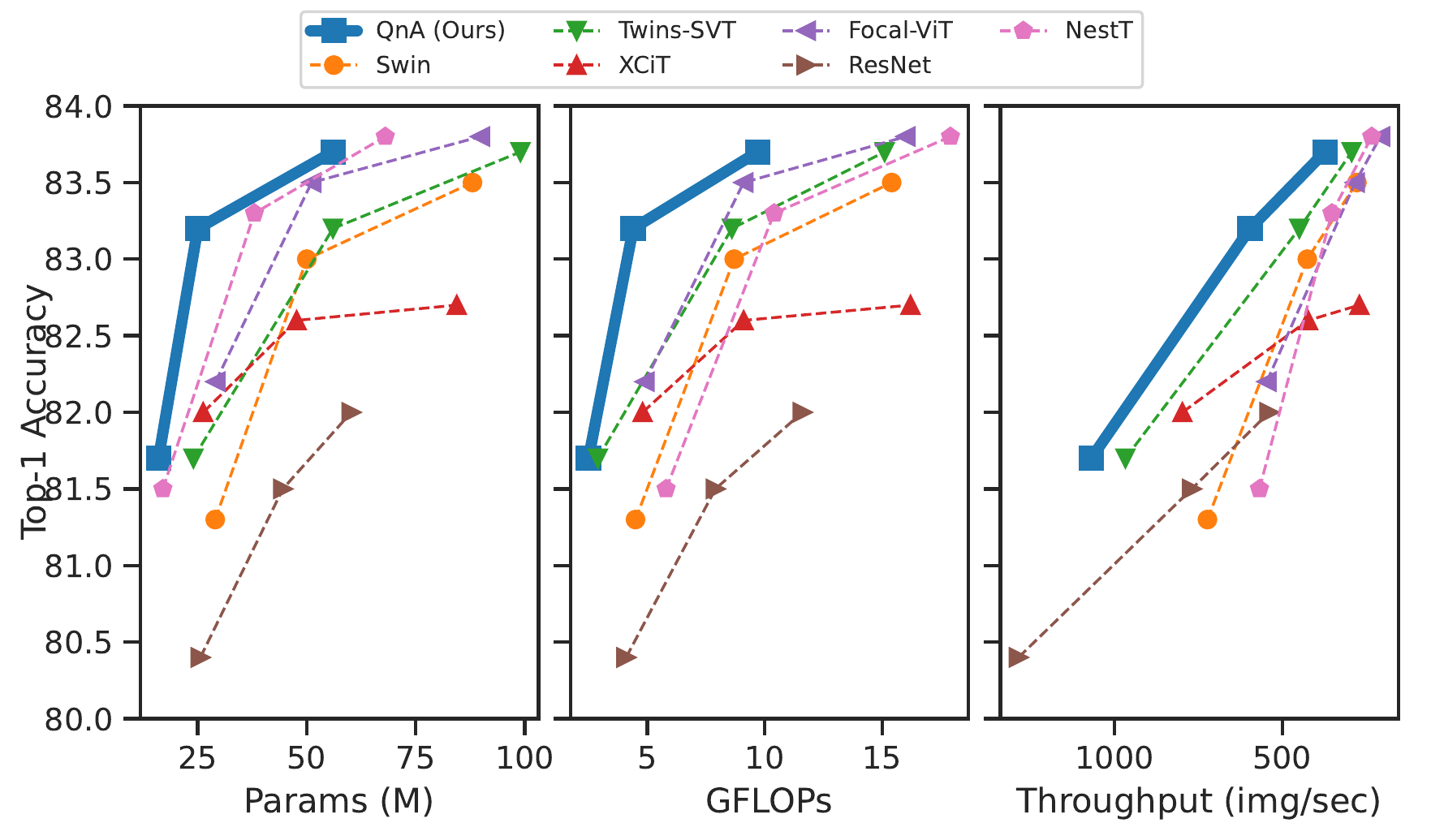}
    \ifarxiv\else
    \vspace{-5pt}
    \fi
    \caption{QnA demonstrates better accuracy-efficiency trade-off compared to state-of-the-art baselines (extracted from the QnA paper).}
    \label{fig:qna_teaser1}
\end{figure}

\ifarxiv
\else
\section{Motion Representation -- Additional Details}
This section completes the Motion Representation section in the main paper.

In this section, we describe the dynamic features predicted by our network.

Recall that $N$ denotes the number of frames in the sequence, and $F$ denotes the length of the features of all joints together in a single motion frame.

\fi %

\section{Hyperparameters and Training Details}  \label{app:hyper_params}
In Tab.~\ref{tab:hyperparams} we detail the values of the hyperparameters that have been used to produce the results shown in this work. Our models have been trained on an NVIDIA GeForce RTX 2080 Ti GPU.

\section{QnA Recap} \label{app:qna}

QnA layers~\cite{arar2022learned} are a fundamental component in our suggested architecture. In this section, we provide an overview of its underlying implementation and illustrate it in Fig.~\ref{fig:qna_overview}. In particular, QnA is an efficient attention-based layer, which operates in a shift-invariant manner. For every $k$-size window, the output is calculated using the self-attention mechanism which is commonly used in the transformer architecture~\cite{vaswani2017attention}. The self-attention is calculated by first projecting the input features into keys $K=X W_K$, values $V=X W_V$, and queries $Q=X W_Q$ via three linear projection matrices $W_K,  W_V, W_Q \in \mathbb{R}^{D\times D}$. Then, the output of the self-attention operation is defined by:

\begin{equation}
\label{eq:self-attention}
    \begin{split}
    \text{\textbf{SA}}(X) & = \text{\textbf{Attention}}\left(Q,K\right) \cdot V \\ 
          & = \text{\textbf{Softmax}}\left(Q K^{T} / {\sqrt{D}} \right) \cdot V.
    \end{split}
\end{equation} 

Instead of performing the pricey query-key operation, QnA detours from extracting the queries from the window itself and directly learns them for the whole-training data (see Fig.~\ref{subfig:qna_overview_ours}). Learning the queries preserves the expressive capability of the self-attention mechanism and enables an efficient implementation that relies on simple and fast operations. In particular, a single query $\tilde{q}$ is learned, and the attention is applied locally for every $k$-size window. Therefore, the output at entry $z_i$ becomes: 

\begin{equation}
\label{eq:single_qna}
    z_{i}  = \textbf{Attention}\left(\tilde{q},K_{\mathcal{N}_{i}}\right) \cdot V_{\mathcal{N}_{i}},
\end{equation} where $\mathcal{N}_{i}$ is the set $k$-neighbourhood of frame $i$.

QnA exhibits state-of-the-art accuracy-efficiency trade-off, as depicted in Fig.~\ref{fig:qna_teaser1}.

\section{User Study -- Screenshots}  \label{app:ustudy}
Our user study displays several video clips on each screen, requesting the user to select the one that is more suitable to the examined attribute, which is either  quality, fidelity, or diversity. Screenshots from a representative video for each attribute are shown in Fig.~\ref{fig:user_study_screenshots}.

\clearpage
\begin{table}[htp]
    \caption{Our choice of hyperparameters, given with the same names as used in the code.}
    \vspace{-10pt}
    \centering
    \begin{tabular}{ l | c }
        Name &  Value \\
        \midrule
        
         \underline{UNet related} & \\
         \quad num\_channels & 256 \\
         \quad channel\_mult & 1 \\
         \quad num\_res\_blocks & 1 \\
         \quad kernel\_size & 3 \\
         \quad use\_scalse\_shift\_norm & True \\
         \quad use\_checkpoint & True \\
         \quad use\_attention & True \\
         \quad use\_qna & True \\

         \underline{QnA related} & \\
         \quad head\_dim & 32 \\
         \quad num\_heads & 4 \\

         \underline{Diffusion related} & \\
         \quad diffusion\_steps & 1000 \\
         \quad noise\_schedule & cosine \\

         \underline{Training related} & \\
         \quad batch\_size & 64 \\
         \quad dropout & 0.5 \\
         \quad lr\_method & ExponentialLR \\
         \quad lr\_gamma & 0.99998 \\
         \quad num\_steps & 60000 \\
         \quad padding\_mode & zeros \\
         \quad warmup\_steps & 0 \\
         \quad weight\_decay & 0 \\
        
        \bottomrule
    \end{tabular}
    \label{tab:hyperparams}
    \ifarxiv
    \else
    \vspace{-5pt}
    \fi
\end{table}

\begin{figure}[t]
\centering
    \begin{subfigure}{\linewidth}
    \centering
    \includegraphics[height=.29\textheight,keepaspectratio]{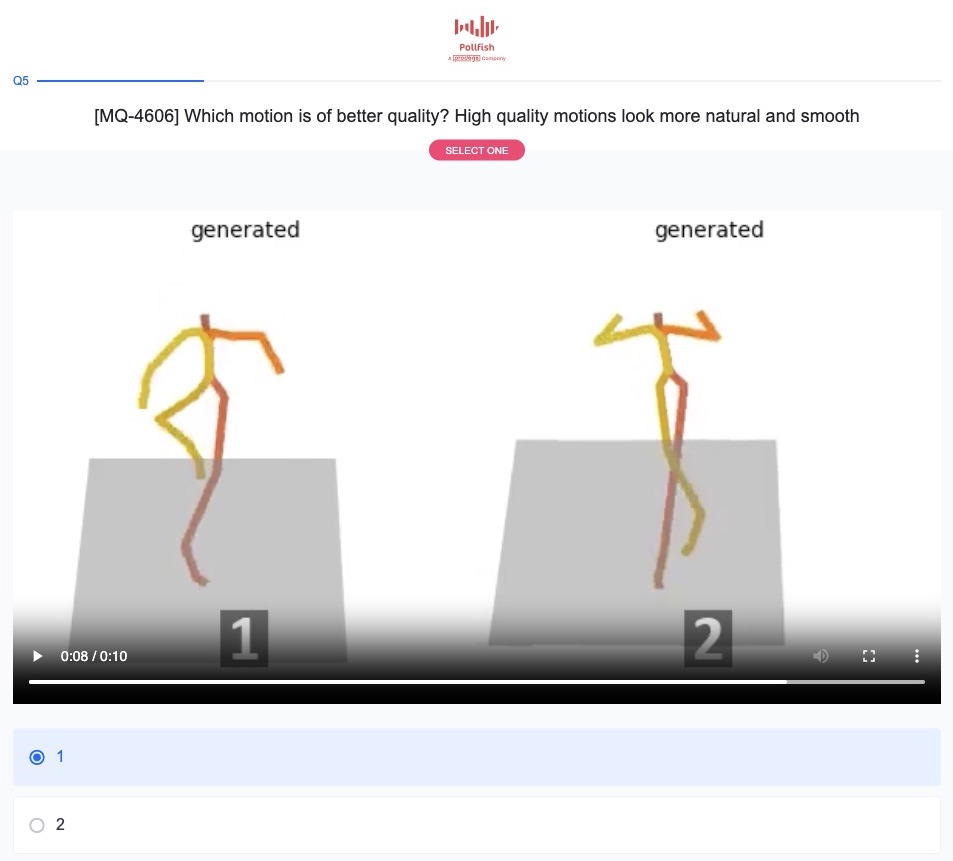} 
\vspace{-5pt}
    \caption{Quality.}
       \end{subfigure}
       
     \vspace{5pt}
    \begin{subfigure}{\linewidth}
    \centering
    \includegraphics[height=.29\textheight,keepaspectratio]{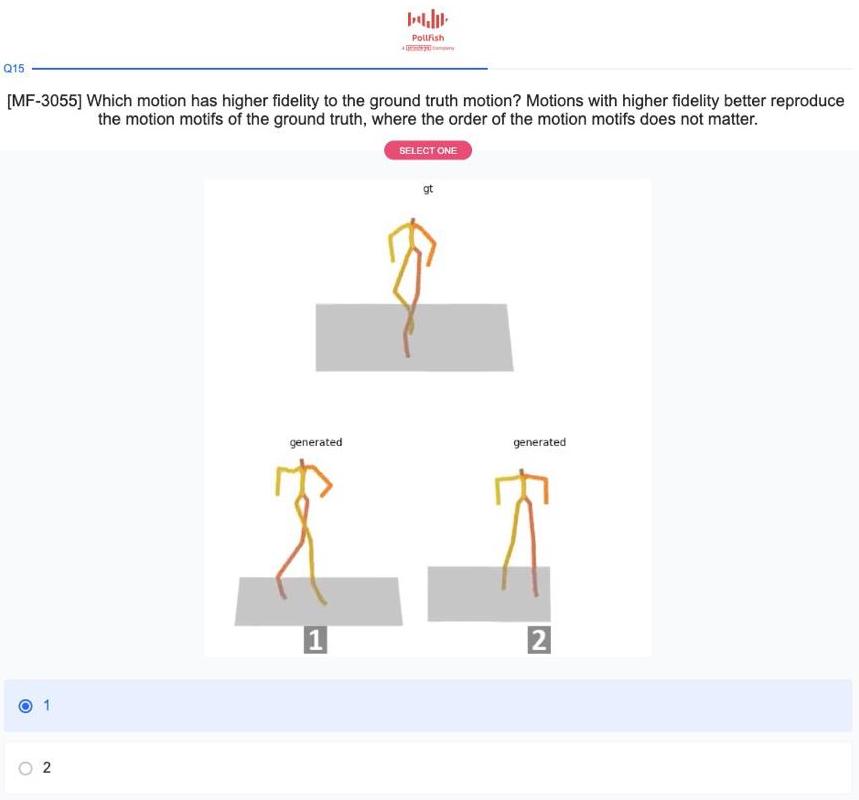} 
\vspace{-5pt}
    \caption{Fidelity.}
    \end{subfigure}
\vspace{5pt}

   \begin{subfigure}{\linewidth}
    \centering
    \includegraphics[height=.29\textheight,keepaspectratio]{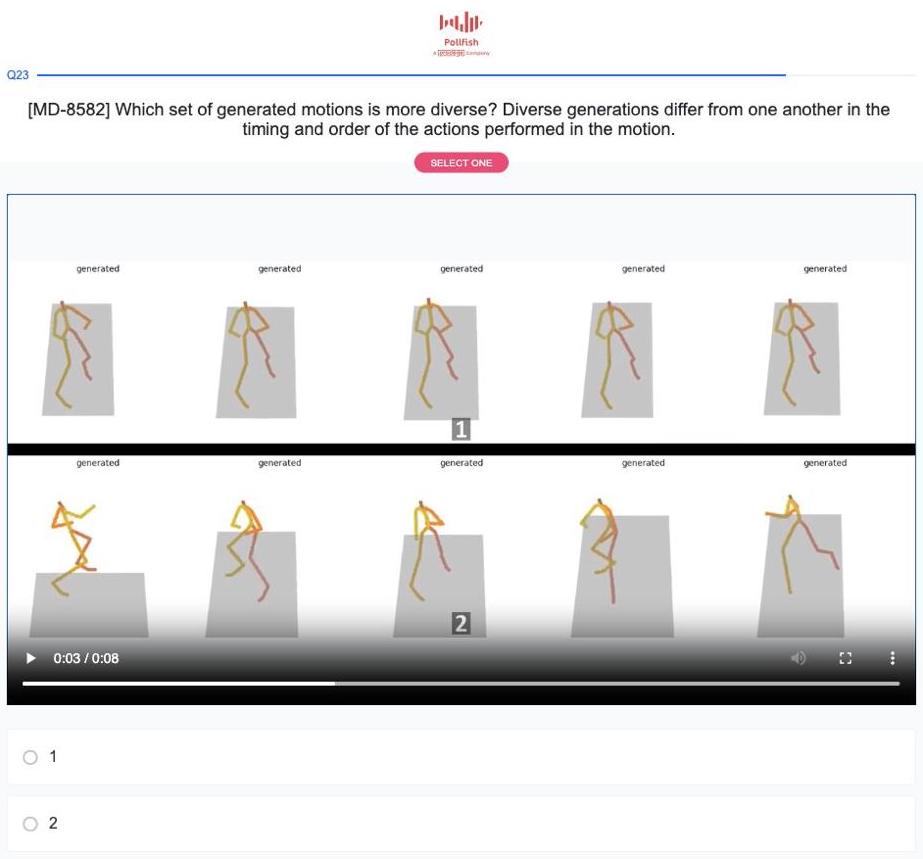}
\vspace{-5pt}
    \caption{Diversity.}
    \end{subfigure}
\hfill
\vspace{-20pt}
\caption{Screenshots from our user study. Note that each human figure in the screenshot is played as a video. }
\label{fig:user_study_screenshots}
\end{figure}

\fi

\end{document}